%% file: main.tex
\documentclass[10pt,twocolumn,letterpaper]{article}

\newcommand{\ours}[0]{EgoAllo} 

\usepackage{cvpr}              %

\usepackage[subtle]{savetrees}
\usepackage{pifont}
\usepackage[table,dvipsnames]{xcolor}%

\definecolor{cvprblue}{rgb}{0.21,0.49,0.74}
\usepackage[pagebackref,breaklinks,colorlinks,allcolors=cvprblue]{hyperref}

\def\paperID{6679} %
\def\confName{CVPR}
\def\confYear{2025}

\title{Estimating Body and Hand Motion in an Ego-sensed World}

\newcommand{\authspace}{\hspace{1em}}
\newcommand{\authspacealt}{\hspace{1em}}
\author{
    \vspace{-1.75em}\\
    {
        Brent Yi\textsuperscript{1}
        \authspace{} Vickie Ye\textsuperscript{1}
        \authspace{} Maya Zheng\textsuperscript{1}
        \authspace{} Yunqi Li\textsuperscript{2}
        \authspace{} Lea M\"uller\textsuperscript{1}
    } \\[0.25em]
    {
        Georgios Pavlakos\textsuperscript{3}
        \authspacealt Yi Ma\textsuperscript{1}
        \authspacealt Jitendra Malik\textsuperscript{1}
        \authspacealt Angjoo Kanazawa\textsuperscript{1}
    } \\[0.875em]
    {
        \textsuperscript{1}UC Berkeley
        \hspace{0.75em}
        \textsuperscript{2}ShanghaiTech
        \hspace{0.75em}
        \textsuperscript{3}UT Austin
    }
}

\begin{document}

\twocolumn[{%
    \renewcommand\twocolumn[1][]{#1}%
    \maketitle
    \centering
    \vspace{-10pt}
    \includegraphics[width=\linewidth]{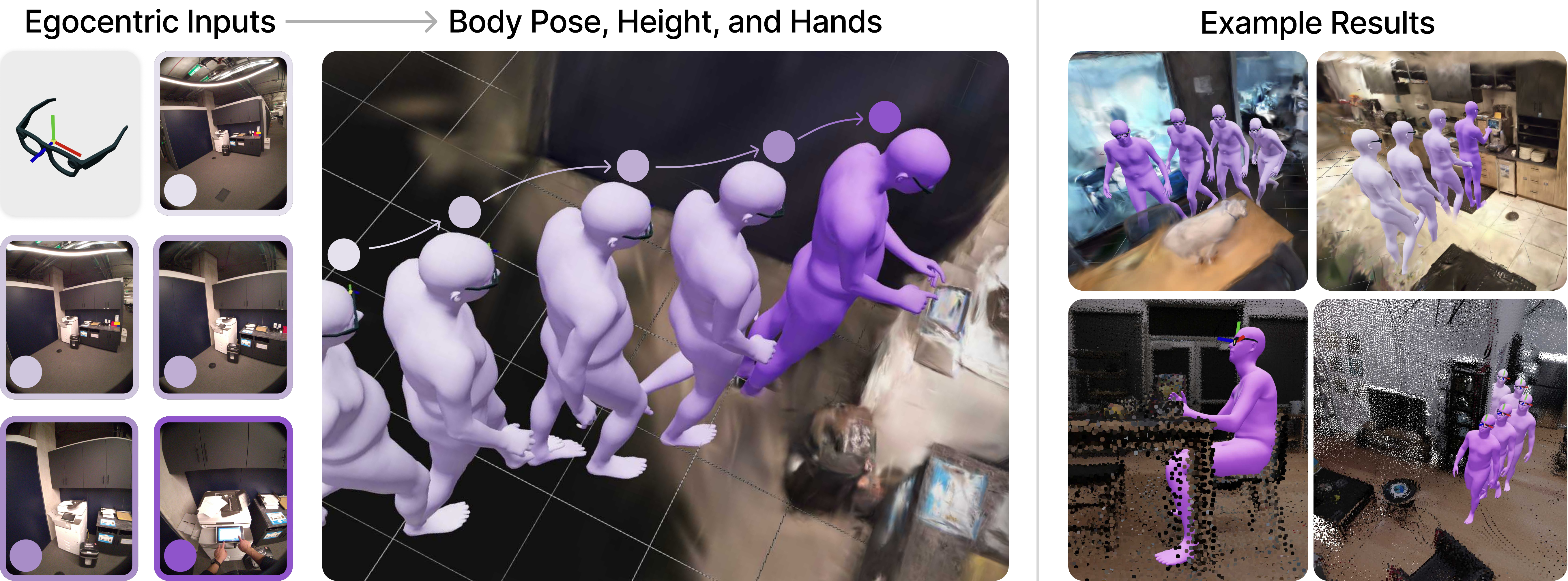}
    \captionsetup{hypcap=false} 
    \captionof{figure}{
        \label{fig:teaser}
        \textbf{EgoAllo.}
        We present a system that estimates human body pose, height, and hand parameters from egocentric SLAM poses and images.
        Outputs capture the wearer's actions in the allocentric reference frame of the scene, which we visualize here with 3D reconstructions.
    }%
    \captionsetup{hypcap=true}
    \vspace{2em}
}]

\input{sec/0_abstract}    
\input{sec/1_intro}

\input{sec/2_related_work}
\input{sec/3_method}

\input{sec/4_experiments}

\input{sec/5_conclusion}

\small \bibliographystyle{ieeenat_fullname} \bibliography{main}

\clearpage
\input{sec/X_suppl}

\end{document}

%% file: sec/0_abstract.tex
\begin{abstract}

We present \ours{}, a system for human motion estimation from a head-mounted device.
Using only egocentric SLAM poses and images, \ours{} guides sampling from a conditional diffusion model to estimate 3D body pose, height, and hand parameters that capture a device wearer's actions in the allocentric coordinate frame of the scene.
To achieve this, our key insight is in representation: we propose spatial and temporal invariance criteria for improving model performance, from which we derive a head motion conditioning parameterization that improves estimation by up to 18\%.
We also show how the bodies estimated by our system can improve hand estimation: the resulting kinematic and temporal constraints can reduce world-frame errors in single-frame estimates by 40\%.

\end{abstract}

%% file: sec/1_intro.tex
\section{Introduction}
\label{sec:intro}

Head-mounted devices are becoming increasingly mainstream. %
In addition to offering new challenges for 3D scene understanding~\cite{zhang2024egogaussian,gu2024egolifter,plizzari2024spatial,mai2023egoloc}, egocentric sensors from these devices are unique in that their outputs are coupled to a human wearer's motion in the world.
Using these sensors to understand the wearer \textit{in addition} to the scene around them is essential for applications in augmented reality, robotics, and assistive technologies.

We therefore introduce EgoAllo, a system that uses egocentric inputs to estimate the wearer and their motion in the world, or allocentric, coordinate frame.
We take as input sensed metric SLAM head poses and egocentric video from devices like Project Aria~\cite{pan2023aria}.
We then estimate as output human body pose, height, and hand motion parameters.

This is a difficult task: while body parts like hands occasionally appear in egocentric frames, most body parameters are never directly observed.
To ensure that estimates are consistent with both the scene and sensed egomotion, harmony is also required between pose \textit{and} height parameters. %
This setting differs from most prior works in egocentric human motion estimation~\cite{li2023egoego,castillo2023bodiffusion}, which focus on body pose and do not address the challenges of height and hand motion.

Our proposed system uses a head pose-conditioned diffusion model as a motion prior, as well as a Levenberg-Marquardt guidance optimizer for sampling hand-body sequence that align with image observations. %
Our results are enabled by a key insight: that the representation used for head pose conditioning is critical for accurate full-body motion estimation. %
We study choices for this representation by (1) identifying desirable spatial and temporal invariance properties that are not fulfilled by existing systems and (2)~using these properties to derive improved parameterizations for our motion prior. 

We systematically evaluate our system on four datasets.
For body estimation, we find that improving the conditioning parameterization leads to an accuracy improvement between 4.9\% and 17.9\%.
Furthermore, we observe that the resulting system can improve hand estimation, reducing world-frame errors by over 40\% compared to single-frame estimates.
Code, model, and more results can be found on our \href{https://egoallo.github.io/}{project webpage}.

%% file: sec/2_related_work.tex
\section{Related Work}
\label{sec:relwork}
\noindent
\textbf{3D human recovery from external visual inputs.}
A large body of work has addressed estimating the parameters of human body models like SCAPE~\cite{anguelov2005scape} or SMPL and its variants ~\cite{loper2023smpl,romero2022mano,pavlakos2019smplx} from third-person visual inputs, where human subjects are observed from the view of outside cameras.
The majority of these works focus on extracting 3D representations from single images, for example by lifting 2D keypoint observations to 3D~\cite{martinez2017simple}, via end-to-end regression~\cite{kanazawa2018hmr,kocabas2021pare,omran2018neural,guler2019holopose,rogez2017lcr,joo2021exemplar,pavlakos2018learning}, via optimization~\cite{pavlakos2019smplx,guan2009estimating,lassner2017unite}, or by exploiting synergies between regression and optimization~\cite{kolotouros2019spin}.
When multiple frames are available in the form of a video, temporal context and tracking can also be incorporated~\cite{rajasegaran2022tracking,goel20234dhumans,kanazawa2019hmmr,yuan2022glamr,kocabas2020vibe,pavlakos2022multishot,pavlakos2022friends}.
The inputs (images) and outputs (human meshes) of many of these systems are similar to the egocentric setting addressed by \ours{}, but egocentric devices present unique challenges because the body being estimated is typically \textit{behind} the outwards-facing cameras used as input.

\vspace{0.5em}
\noindent
\textbf{Priors for human motion.}
The primary challenge of ego-sensed human motion estimation is limited observability; a prior is required to resolve ambiguities.
For human motion, these priors are typically framed as unconditional distributions over plausible human motion.
Distributions can be represented either by modeling the physical constraints of our world~\cite{rempe2020contact,li2019estimating,peng2018sfv,brubaker2010physics} or by learning generative models of human motion directly from data.
For learning unconditional priors, classical data-driven approaches include fitting mixtures-of-Gaussians to 3D keypoint trajectories~\cite{howe1999bayesian}, while modern approaches include training variational autoencoders~\cite{kingma2013vae,rezende2014stochastic} to model either autoregressive transitions~\cite{rempe2021humor,ling2020character,ghorbani2020probabilistic} or full spatiotemporal sequences~\cite{he2022nemf}.
After training, these priors can be applied to estimation problems in iterative optimization frameworks~\cite{ye2023slahmr,kocabas2024pace,rempe2021humor}.
\ours{} is built on the same intuition as these methods, but follows previous work in ego-sensed motion estimation and uses a task-specific conditional prior.

\input{figtex/method}

\vspace{0.5em}
\noindent
\textbf{Denoising diffusion for human motion.}
The core of EgoAllo is a denoising diffusion model~\cite{sohl2015deep,ho2020denoising,po2023diffusionsurvey} from which we can sample 3D human body motion. %
While diffusion models are primarily known for their success in text-conditioned image generation~\cite{rombach2022latentdiffusion,saharia2022imagen}, they have also enabled advances in human motion synthesis conditioned on modalities like text~\cite{kim2023flame,zhang2022motiondiffuse,karunratanakul2023gmd}, music~\cite{tseng2023edge,alexanderson2023listen}, poses~\cite{li2023egoego,karunratanakul2023gmd}, and object geometry~\cite{kulkarni2023nifty,li2023controllable, li2023object}.
\ours{} adopts a similar conditional diffusion approach, while specifically studying the design of conditioning parameters used for ego-sensed human motion estimation. 
The iterative nature of denoising diffusion also enables guidance~\cite{dhariwal2021diffusionbeats,song2020score,karunratanakul2023gmd,zhang2023probabilistic,jiang2024back,ci2023gfpose}, where denoising steps are steered to satisfy a desired objective.
We use guidance to incorporate observations like visual hand pose observations during test-time.  %

\vspace{0.5em}
\noindent
\textbf{Human motion from egocentric observations.}
\ours{} builds on intuition from several prior works in egocentric sensing for human motion estimation.
Many rely on fisheye cameras that place the wearer's body into the field of view~\cite{jiang2021egocentric,tome2020selfpose,tome2019xr,wang2021estimating,xu2019mo,rhodin2016egocap,tome2019xr,wang2024egocentric}.
Other approaches rely on body-mounted cameras~\cite{shiratori2011motion}, simulation-based physical plausibility~\cite{yuan20183d,yuan2019ego,luo2021dynamics}, body- and hand-mounted inertial sensors~\cite{yi2021transpose,yi2023egolocate,lee2024mocap,kim2022fusion}, handheld controllers~\cite{castillo2023bodiffusion,jiang2022avatarposer,jiang2023egoposer}, and interaction cues from other humans~\cite{ng2020you2me}.
Concurrent works have also used the Nymeria~\cite{ma2024nymeria} 
dataset for egocentric motion with language description outputs~\cite{hong2024egolm}, as well for online settings with scene geometry and CLIP~\cite{radford2021learning} feature inputs~\cite{guzov2024hmd}.
Most relevantly, EgoEgo~\cite{li2023egoego} demonstrates how human body poses can be estimated offline without body observability assumptions.
The authors accomplish this by carefully integrating several components: a monocular SLAM system~\cite{teed2021droid}, a pose-conditioned gravity vector regression network, an optical flow feature-conditioned head orientation and scale regression network, and a head pose-conditioned body diffusion model.
\ours{} differs in both inputs---we study conditioning parameters computed from the metric SLAM poses provided by devices like Project Aria~\cite{somasundaram2023projectaria}---and outputs---we consider body height variation and hand poses.

\vspace{0.5em}
\noindent
\textbf{Conditioning for ego-sensed poses.}
Prior works vary in how head pose information is parameterized and used as neural network input.
AvatarPoser~\cite{jiang2022avatarposer} and BoDiffusion~\cite{castillo2023bodiffusion} parameterize head pose as four components: world-frame orientation, orientation deltas, world-frame position, and world-frame position deltas. These works focus on settings with VR controller input, and parameterize controller pose inputs the same way.
EgoEgo~\cite{li2023egoego}'s diffusion model uses only absolute head positions and orientations, but similar to HuMoR~\cite{rempe2021humor}, in implementation defines a per-sequence canonical coordinate frame to ensure that all input trajectories passed to the model are aligned with the same initial $xy$ position and forward direction.
In our work, we refer to this as \textit{sequence canonicalization}.
Finally, EgoPoser~\cite{jiang2023egoposer} proposes a similar scheme that aligns initial positions for both head pose and controller pose inputs.
We propose an alternative to these parameterizations that is motivated by the robustness and generalization benefits of invariance, as observed in prior work for designing both representations~\cite{wiskott2002slow,lowe2004distinctive,savarese20073d,chen2020simclr,zhang2012tilt,yi2023tilted,zhan2023general} and neural network architectures~\cite{lecun1995convolutional,kanazawa2014locally,zaheer2017deep,thomas2018tensor,chen2021equivariant,charles2017pointnet,cohen2016group,feng2023generalizing,yang2023equivact,yang2024equibot}.
Specifically, we introduce in Section~\ref{sec:invariant_conditioning} a parameterization that is invariant to both spatial and temporal shifts.

%% file: figtex/method.tex
\begin{figure*}[t!]
  \centering
  \includegraphics[width=\linewidth]{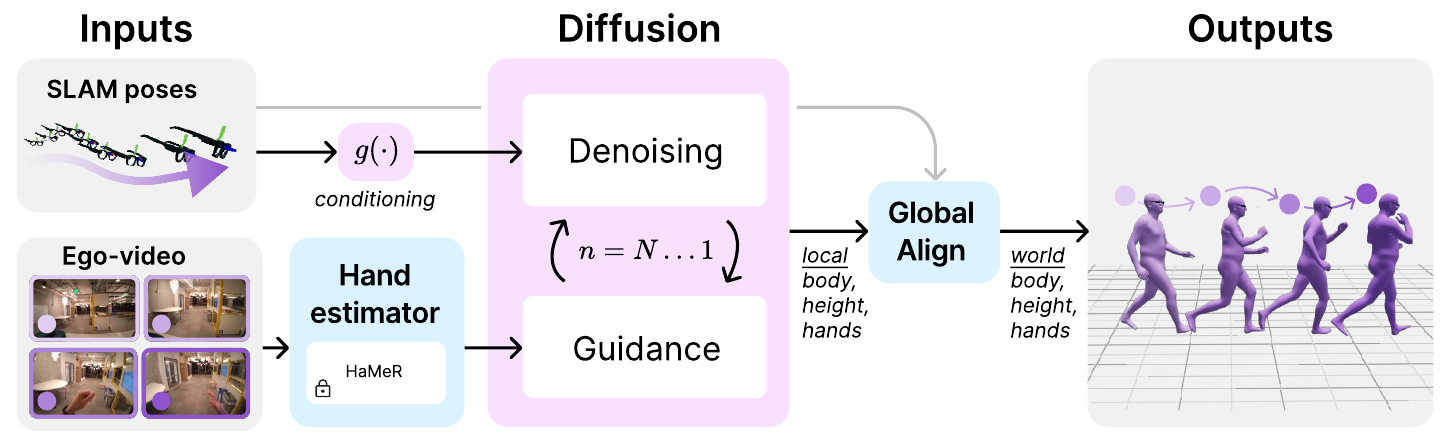}
  \caption{
    \textbf{Overview of components in \ours{}.}
    We restrict the diffusion model to local body parameters (Section~\ref{sec:diffusion_output_representation}).
    An invariant parameterization $g(\cdot)$ (Section~\ref{sec:invariant_conditioning}) of SLAM poses is used to condition a diffusion model.
    These can be placed into the global coordinate frame via global alignment (Section~\ref{sec:global_alignment}) to input poses.
    When available, egocentric video is used for hand detection via HaMeR~\cite{pavlakos2023reconstructing}, which can be incorporated into samples via guidance (Section~\ref{sec:guidance_losses}).
  }
  \label{fig:method}
\end{figure*}

%% file: sec/3_method.tex
\newcommand{\mb}[1]{\mathbb{#1}}
\newcommand{\tf}[1]{\mathbf{#1}}
\newcommand{\mc}[1]{\mathcal{#1}}
\newcommand{\fw}{{}^{\textrm{w}}}
\newcommand{\fp}{{}^{\textrm{p}}}
\newcommand{\fc}{{}^{\textrm{c}}}
\newcommand{\ps}[1]{{}^{\textrm{#1}}}
\newcommand{\mn}[1]{{#1}^m_n}
\newcommand{\trm}[1]{\textrm{#1}}

\section{Method}
\label{sec:method}

We study the problem of using sensors from an egocentric device to estimate the actions of the wearer in an allocentric coordinate frame.
We assume a flat floor and two inputs: poses from the device's SLAM system and egocentric video.

Our system uses head pose information to condition a diffusion-based prior over body pose and height, and incorporates visual hand observations during sampling.
This allows it to benefit from both 3D human motion capture datasets~\cite{mahmood2019amass}, which are used for the motion prior, and from large-scale image datasets~\cite{pavlakos2023reconstructing}, which are used for hand estimates.

\subsection{Ego-conditioned motion diffusion}
\label{sec:method_diffusion_model}

\textit{Notation:} we use $\mathbf{T}_{\text{A}, \text{B}} = (\mathbf{R}_{\text{A}, \text{B}}, \mathbf{p}_{\text{A}, \text{B}})$ to denote an SE(3) transform to frame A from frame B, composed of rotation ($\mathbf{R}_{\text{A},\text{B}}$) and position ($\mathbf{p}_{\text{A},\text{B}}$) terms.
Temporal steps $t$ are superscripted and diffusion noise steps $n$ are subscripted.
$\vec{x}_0^t$ thus refers to the $t$-th timestep of a clean ($n=0$) human motion sequence.

Given an observation window of $T$ timesteps, \ours{}'s motion prior is a diffusion model that aims to capture the distribution of human motions $\vec{x}_0 = \{\vec{x}_0^{\ 1}, \dots, \vec{x}_0^{\ T}\}$
conditioned on head pose encodings $\vec{c} = \{\vec{c}^{\ 1}, \dots, \vec{c}^{\ T}\}$.
For each timestep $t$, we represent human motion in the form of SMPL-H~\cite{loper2023smpl,romero2022mano} model parameters $\{\mathbf{T}^t_{\text{world}, \text{root}}, \Theta^t, \beta\}$:
root transforms $\mathbf{T}^t_{\text{world}, \text{root}} \in \text{SE}(3)$, where the person's root frame is located at their pelvis, local joint rotation matrices $\Theta^t \in \mathbb{R}^{51 \times 3 \times 3}$, and time-invariant shape $\beta \in \mathbb{R}^{16}$.

Dependencies between local joint rotations, body size variation, and global motion make this learning task a challenging one.
Our key insight is that this difficulty can be reduced by designing parameterizations with desirable invariance properties.
Spatial and temporal invariances allow the model to focus on the essential structure of motion, without being affected by irrelevant shifts in position or time.

\subsubsection{Diffusion output representation}
\label{sec:diffusion_output_representation}
As output, we sample body and hand joint rotations, body shapes, and binary contact predictions $\vec{x}_0^t = \{\Theta^t, \beta^t, \psi_{j=1\dots21}^t\}$, where body shape $\beta^t$ is supervised to be equal for all timesteps and $\psi_j^t$ is a per-joint contact indicator.
Notably, these parameters are all local---we discuss how outputs can be placed into the allocentric coordinate frame in Section~\ref{sec:global_alignment}.

We choose this output set for three main reasons.
(1)~Body shape encodes the wearer's height, which is critical for grounding in the metric-scale geometry of the scene.
This is rarely considered by prior work: with the exception of \cite{jiang2023egoposer}, which is focused on tracking with controller input, existing methods~\cite{li2023egoego,castillo2023bodiffusion,jiang2022avatarposer} otherwise produce outputs using a fixed ``mean'' human shape.
(2)~Contact predictions enable losses for common problems like foot skating, which are discussed in Section~\ref{sec:guidance_losses}.
(3)~Finally, local bodies are invariant to the global coordinate frame.
As we discuss next, the conditioning parameterization for the model can therefore also be invariant to arbitrary transformations along the floor plane.

\subsubsection{Invariant conditioning}
\label{sec:invariant_conditioning}

The goal of our conditioning representation is to map raw SLAM poses (head motion) to a parameterization that is amenable to learning for the diffusion model. %

\vspace{0.5em}
\noindent
\textbf{Raw inputs.}
To capture the head motion at each time step, we assume as input poses of a \textit{central pupil frame} (CPF), which the SLAM systems of devices like Project Aria can provide with millimeter-level accuracy~\cite{somasundaram2023projectaria}.
For time $1\dots T$, we reparameterize these poses for conditioning using a function $g$: %
\begin{align}
    \mathbf{T}_{\text{world},\text{cpf}}^t &= (\mathbf{R}_{\text{world},\text{cpf}}^t,\mathbf{p}_{\text{world},\text{cpf}}^t) \in \text{SE}(3),\\
    \{\vec{c}^{\ 1}, \dots, \vec{c}^{\ T}\} &= g(\{\mathbf{T}_{\text{world},\text{cpf}}^1, \dots, \mathbf{T}_{\text{world},\text{cpf}}^T\}).\label{eq:g_definition}
\end{align}
The CPF frame differs from prior works that condition on a coordinate frame attached to the SMPL human model's ``head joint''~\cite{li2023egoego,castillo2023bodiffusion,jiang2022avatarposer,jiang2023egoposer}.
The offset between this head joint and the device pose depends on the head shape captured by $\beta^t$, and is thus difficult to precompute in our setting.

To encode absolute height, we assume that the world frame's +$z$-axis faces upwards, and that the ground is located at $z=0$.
Ground parameters are directly available in the training data~\cite{mahmood2019amass}; at test time, we can also extract these parameters from sparse SLAM points via RANSAC (Appendix~\ref{app:floor_height}).

\vspace{0.5em}
\noindent
\textbf{Invariance goals.}
As discussed in Section~\ref{sec:relwork}, prior work varies in how the function $g$ is implemented.
To understand how choices impact learning, we propose two invariance properties for head motion representations.
Each reduces representational redundancy, which eases the learning problem. %
\newtheorem{invariance}{Invariance}
\begin{invariance}[Spatial]
    \label{inv:spatial_invariance}
    Global transformations along the floor plane should not affect a person's local motion.
    Given $\mathbf{T}_\text{xy} \in$ SE(3) restricted to the XY plane, $g$ should fulfill
    $
        g(\{\mathbf{T}_\text{xy}\mathbf{T}_{\text{world},\text{cpf}}^t\}_t) = g(\{\mathbf{T}_{\text{world},\text{cpf}}^t\}_t)
        \ \forall\ \mathbf{T}_\text{xy}
    $.
\end{invariance}
\begin{invariance}[Temporal]
    \label{inv:temporal_invariance}
    Head motion representations for a given body motion should be independent of location within a temporal window.
    This can be expressed as temporal shift equivariance.
    Let $\vec{c}^{\ t}$ be as defined in Equation~\ref{eq:g_definition}.
    For any shift $\delta$ such that
    $\{\vec{c}^{\ 1}_\text{shift},\dots,\vec{c}^{\ T}_\text{shift}\} = g(\{\mathbf{T}_{\text{world},\text{cpf}}^{1+\delta}, \dots, \mathbf{T}_{\text{world},\text{cpf}}^{T+\delta}\})$,
    $g$ should satisfy $\vec{c}^{\ t}_\text{shift} = \vec{c}^{\ t + \delta}$ for overlapping timesteps.
\end{invariance}
No parameterization used by existing work satisifies both of these properties.
The sequence canonicalization approach of EgoEgo~\cite{li2023egoego} achieves spatial invariance (Invariance~\ref{inv:spatial_invariance}), but inserts a sequence-wide dependency on the first timestep of each window that results in a violation of Invariance~\ref{inv:temporal_invariance}.
The absolute poses and pose deltas used by~\cite{castillo2023bodiffusion,jiang2022avatarposer} satisfy Invariance~\ref{inv:temporal_invariance}, but not Invariance~\ref{inv:spatial_invariance}.
Finally, the relative positions considered by~\cite{jiang2023egoposer} are neither spatially nor temporally invariant.

\begin{figure}[t!]
  \vspace{-1em}
  \centering
    \includegraphics[width=0.75\linewidth]{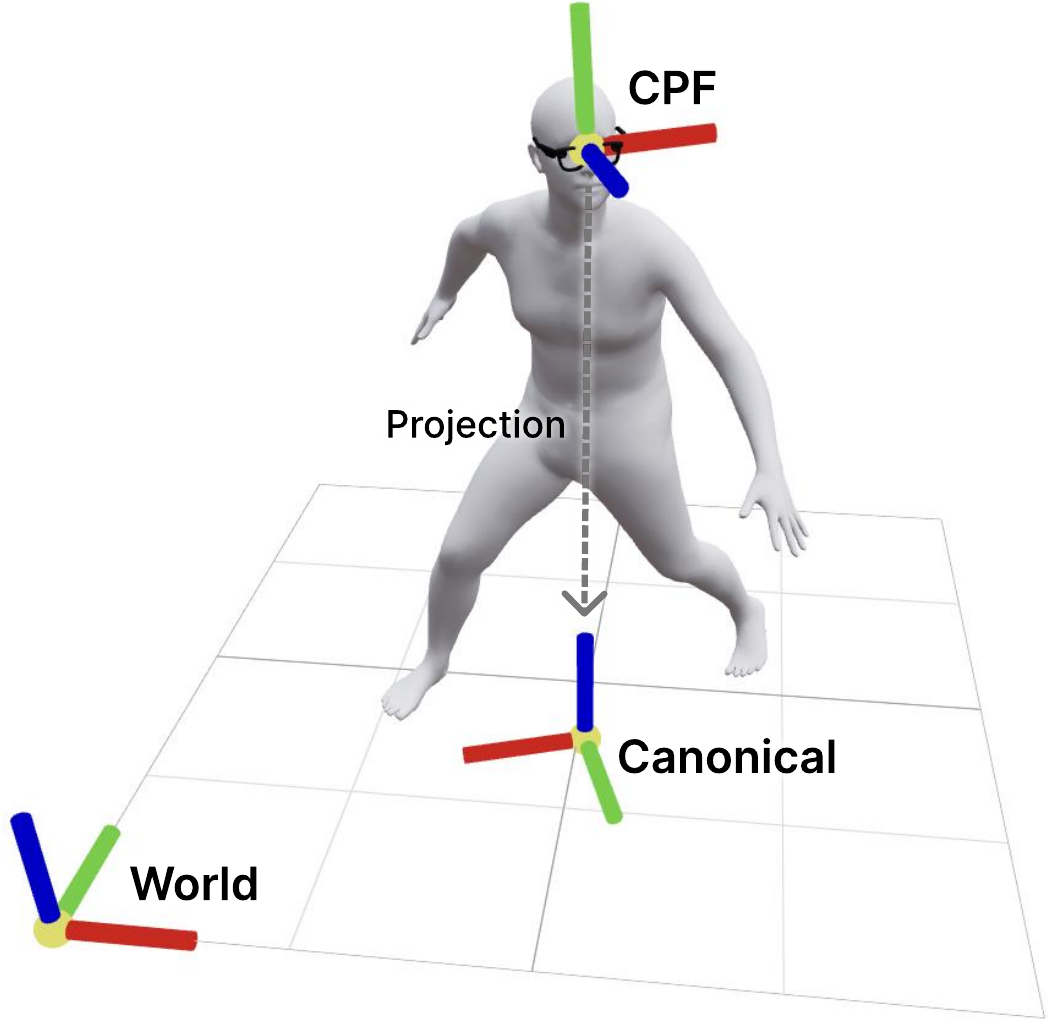}
  \caption{
    \textbf{Locally canonicalized coordinate frames.}
    We compute our invariant conditioning parameterization (Equation~\ref{eq:cond_param}) using transformations computed from three coordinate frames.
    Following \cite{somasundaram2023projectaria}, the CPF has the $z$-axis forward.
    Following HuMoR~\cite{rempe2021humor}, the world and canonical $z$-axes point up.
    Canonical frames are computed by projecting the CPF frame origin to the ground plane, then aligning the canonical $y$-axis to the CPF forward direction.
  }
  \label{fig:coordinate_frames}
  \vspace{-1em}
\end{figure}

\vspace{0.5em}
\noindent
\textbf{Invariant conditioning.}
We propose a formulation for $g$ that achieves both invariance properties by locally canonicalizing head motion with respect to the floor at each timestep.
We build on the relative motion of the CPF frame at each time $t$, which respects both Invariance~\ref{inv:spatial_invariance} and~\ref{inv:temporal_invariance}:
\begin{align}
    \Delta\mathbf{T}_{\text{cpf}}^{t-1, t} &= (\mathbf{T}_{\text{world},\text{cpf}}^{t-1})^{-1}\mathbf{T}_{\text{world},\text{cpf}}^t.
    \label{eq:relative_cpf_motion}
\end{align}
Importantly, the translation component of this transformation is in the local frame.
This is distinct from world-frame position deltas~\cite{jiang2022avatarposer,castillo2023bodiffusion,jiang2023egoposer}, which still violate Invariance~\ref{inv:spatial_invariance}.

Relative transforms alone do not encode information relative to the scene or floor: full trajectories can even be flipped upside down without impacting $\Delta \mathbf{T}_\text{cpf}^{t-1,t}$.
We therefore propose to ground relative motion to the floor plane with a transformation between the CPF frame and a \textit{per-timestep} canonical frame, which is computed by projecting the CPF frame to the floor.
This encodes head height and orientation.
Our full representation then becomes:
\begin{align}
    \label{eq:cond_param}
    \vec{c}^{\ t} &= \underbrace{
    \left\{
    \Delta\mathbf{T}_{\text{cpf}}^{t-1, t}, \quad
    (\mathbf{T}_{\text{world},\text{canonical}}^t)^{-1}
    \mathbf{T}_{\text{world},\text{cpf}}^t
    \right\}}_{\text{Invariant implementation of}\ g(\cdot)}.
\end{align}
We visualize an example of a canonical frame in Figure~\ref{fig:coordinate_frames} and our full representation in Appendix~\ref{app:conditioning_vis}.
Canonical frames are positioned by projecting the CPF origin to the floor plane; given standard bases $\mathbf{e}_{\{x,y,z\}}$, we compute:
\begin{align}
    \mathbf{p}_{\text{world},\text{canonical}}^t &= \begin{bmatrix}
        \:\mathbf{e}_x
        & \mathbf{e}_y
        & \vec{0}
    \end{bmatrix}^\top
    \mathbf{p}_{\text{world},\text{cpf}}^t.
    \label{eq:canonical_position}
\end{align}
For orientation, we align the canonical frame's local $z$-axis parallel to the world $z$-axis and its local $y$-axis toward the ``forward'' direction $\vec{v}^{\:t}$ of the CPF frame.
With $\mathbf{R}_z(\cdot): \mathbb{R} \to \text{SO}(3)$ constructing a $z$-axis rotation and $\mathbf{e}_{\{x,y,z\}}$ again as standard bases, we compute this as:
\begin{align}
    \vec{v}^{\:t} &= \mathbf{R}_{\text{world},\text{cpf}}^t\:\mathbf{e}_z,\\
    \mathbf{R}_{\text{world},\text{canonical}}^t &= \mathbf{R}_z \left(-\text{arctan2}\left(\mathbf{e}_x^\top\vec{v}^{\ t}, \mathbf{e}_y^\top\vec{v}^{\ t}\right)\right). 
    \label{eq:canonical_rotation}
\end{align}
This canonical frame definition is an important departure from prior work.
While EgoEgo~\cite{li2023egoego} and HuMoR~\cite{rempe2021humor} use similar canonical frames, they only compute one per sequence.
Instead, we compute Equations~\ref{eq:canonical_position} and~\ref{eq:canonical_rotation} at every timestep.
This enables floor plane grounding without sacrificing Invariance~\ref{inv:temporal_invariance}.

\subsection{Estimation via sampling}
\label{sec:method_sampling}

We use our local body representation and invariant conditioning strategies to train a motion prior in the form of a denoising diffusion model~\cite{ho2020denoising}.
Given diffusion step $n = N \dots 1$, we follow~\cite{ramesh2022hierarchical} and approximate the denoising process as:
\begin{equation}
    p_\theta(\vec x_{n-1} | \vec x_n, \vec c) = \mc{N}(\mu_\theta(\vec x_n, n, \vec c), \sigma_n^2 \tf I),
\end{equation}
where a transformer~\cite{vaswani2017attention} $\mu_\theta$ is trained to predict the posterior mean from noised sample $\vec x_n$ and conditioning $\vec c$.
With noise-dependent weight term $w_n$, the loss can be written as:
\begin{equation}
    \label{eq:diffusion_training_loss}
    \min_{\theta}\ 
    \ \mathbb{E}_{\vec{x}_0}
    \mathbb{E}_{n \sim \mathcal U} \left[ w_n \lVert \mu_\theta(\vec{x}_n, n, \vec{c}) - \vec{x}_0 \rVert^2 \right].
\end{equation}
After training, we estimate human motions by following DDIM~\cite{song2020denoising} for sampling.
The final \ours{} sampling procedure includes several additional components: a global alignment phase, guidance losses for physical constraints and visual hand observations, and a path fusion~\cite{bar2023multidiffusion} approach for longer sequence lengths.
We describe these below.

\subsubsection{Global alignment}
\label{sec:global_alignment}
To place sampled bodies into the allocentric coordinate system, we compute the absolute pose of the SMPL-H root as:
\begin{align}
\mathbf{T}_{\text{world},\text{root}}^t &=
\mathbf{T}_{\text{world},\text{cpf}}^t
\mathbf{T}_{\text{cpf},\text{root}}^{(\Theta^t, \beta^t)},
\end{align}
where $\mathbf{T}_{\text{cpf},\text{root}}^{(\Theta^t, \beta^t)}$ computes the transform between the root of the human and their CPF frame for a given set of local pose and shape parameters.
Similar processes are applied in~\cite{jiang2023egoposer,jiang2022avatarposer,castillo2023bodiffusion}.
In contrast to directly outputting absolute body transformations from the diffusion model~\cite{li2023egoego}, this guarantees exact alignment between estimates and input SLAM sequences. %

\subsubsection{Guidance losses}
\label{sec:guidance_losses}
Our diffusion model learns a distribution of human motion conditioned on the central pupil frame motion.
At test time, we incorporate constraints from physical priors and visual hand observations via guidance~\cite{zhang2023probabilistic,jiang2024back,ci2023gfpose}.
Similar to~\cite{li2023controllable, karunratanakul2023gmd}, we accomplish this by applying costs to the joint rotations $\Theta = \{\Theta^1, \dots, \Theta^T\}$ predicted by $\mu_\theta(\vec{x}_n, n, \vec{c})$.
We treat the body shape $\beta^t$ and contacts $\psi^t_{j=1\dots21}$ as fixed and optimize over body and finger pose to minimize hand observation, skating, and prior costs with a Levenberg-Marquardt optimizer:
\begin{align}
    \mathcal{E}^{(\Theta)}_\text{guidance} = \mathcal{E}^{(\Theta)}_\text{hands} + \mathcal{E}^{(\Theta)}_\text{skate} + \mathcal{E}^{(\Theta)}_\text{prior}.
\end{align}

We begin by running HaMeR on the egocentric image corresponding to each timestep $t$.
When detected, this produces 3D hand estimates in the form of MANO~\cite{romero2022mano} joint parameters and camera-centric 3D hand keypoints $\hat{\mathbf{p}}^t_\text{camera,j}$ for hand joint set $j \in \mathcal{H}$.
Optionally, wrist and palm poses can also be estimated using Project Aria's Machine Perception Services~\cite{somasundaram2023projectaria}.
With each subcripted $\lambda$ indicating a scalar weighting term, we have:
\begin{align}
    \mathcal{E}_\text{hands}^{(\Theta)} &=   \lambda_\text{hands3D}\mathcal{E}_\text{hands3D}^{(\Theta)} + \lambda_\text{reproj}\mathcal{E}_\text{reproj}^{(\Theta)}.%
\end{align}
The 3D objective $\mathcal{E}_\text{hands3D}^{(\Theta)}$ minimizes the distance between the detected hand parameters and the corresponding SMPL-H hand parameters, in terms of wrist pose and local joint rotations.
With $\Pi_K$ as projection with camera intrinsics $K$, $\mathbf{p}_{world,j}^{(\Theta^t)} \in \mathbb{R}^3$ as the world position for joint $j$ at time $t$, and $\mathbf{T}_{\text{camera},\text{cpf}}$ from the device calibration, the reprojection cost is:
\begin{align}
    \mathcal{E}_\text{reproj}^{(\Theta)} &= \sum_{t, j \in \mathcal{H}} \lvert\lvert
        \Pi_K(\mathbf{p}^{(\Theta^t)}_{\text{camera},j}) - \Pi_K(\hat{\mathbf{p}}^t_{\text{camera},j})
    \rvert\rvert_2^2,\\
    \mathbf{p}^{(\Theta^t)}_{\text{camera},j} &= \mathbf{T}_{\text{camera},\text{cpf}} (\mathbf{T}_{\text{world},\text{cpf}}^t)^{-1}
        \mathbf{p}_{\text{world},j}^{(\Theta^t)}.
    \label{eq:reprojection}
\end{align}
To reduce foot skating, we use contact predictions to apply a skating cost~\cite{rempe2021humor,ye2023slahmr} for each time $t$ and joint $j$:
\begin{align}
    \mathcal{E}_\text{skate}^{(\Theta)} = \sum_{t,j} \lambda_\text{skate}\lvert\lvert \frac{1}{2}(\psi^t_j+ \psi^{t-1}_j)(\mathbf{p}_{world,j}^{t} - \mathbf{p}_{world,j}^{t-1}) \rvert\rvert_2^2.%
\end{align}
Finally, we minimize a prior cost $\mathcal{E}^{(\Theta)}_\text{prior}$.
This cost penalizes deviations between joint rotations $\Theta^t$ and rotations $\hat{\Theta}^t$ from the denoiser $\mu_\theta(\vec{x}_n, n, \vec{c})$.
We include terms for absolute rotation, rotational velocity, and forward kinematics position.
\subsubsection{Sequence length extrapolation}
For longer sequences at test time, we draw on existing methods in compositional generation for both image~\cite{bar2023multidiffusion,zhang2023diffcollage} and human motion~\cite{shafir2023human,barquero2024seamless} diffusion models.
We train our motion prior using subsequences of up to length 128; when input observations exceed this length at test time, we split into windows with a 32-timestep overlap between neighbors.
We then run our model $\mu_\theta(\vec{x}_n, \vec{c}, n)$ on windows in parallel.
Diffusion paths for overlapping regions are fused following MultiDiffusion~\cite{bar2023multidiffusion} after each denoising step.

%% file: sec/4_experiments.tex
\input{figtex/cond_compare}

\section{Experiments}
\label{sec:experiments}

We conduct a series of experiments to evaluate \ours{}'s conditioning parameterization, body estimation accuracy, and hand estimation performance.

\vspace{0.5em}
\noindent
\textbf{Training.}
To train \ours{} models used in our experiments, we need sequences containing human body and hand pose parameters, body shapes, and device SLAM poses $\textbf{T}_{\text{world},\text{cpf}}^t$.
Similar to prior work~\cite{li2023egoego,castillo2023bodiffusion,jiang2022avatarposer}, we train \ours{} using AMASS~\cite{mahmood2019amass} with synthesized device poses.
We annotate train split sequences by anchoring a central pupil frame between vertices corresponding to the left and right pupils in the blend skinned mesh, and at train time sample sequences between length 32 and 128.

\vspace{0.5em}
\noindent
\textbf{Evaluation.}
We evaluate with four datasets.
We use AMASS~\cite{mahmood2019amass}, RICH~\cite{Huang:CVPR:2022}, and Aria Digital Twins (ADT)~\cite{pan2023aria} for body estimation evaluation, and EgoExo4D~\cite{grauman2023egoexo4d} for hand estimation evaluation.
AMASS and RICH do not include egocentric data; we annotate these with synthetic device poses using the same procedure we use for training.
ADT and EgoExo4D both include egocentric images and SLAM poses captured using Project Aria glasses~\cite{somasundaram2023projectaria}, which we use directly.

\vspace{0.5em}
\noindent
\textbf{Metrics.}
To quantify performance, we report four metrics:
\textbf{(1)~MPJPE} is a world-frame mean per-joint position error (millimeters).
\textbf{(2)~PA-MPJPE} is the Procrustes-aligned mean per-joint position error in millimeters, where joint positions are aligned on a per-timestep basis before error are computed.
\textbf{(3)~GND} is a grounding metric, designed in response to a phenomena where ego-sensed humans ``float'' above the ground. Given a human body trajectory, this metric contains a simple binary indicator of whether the feet of the human ever touch the ground plane.
\textbf{(4)}~$\mathbf{T}_\text{head}$ is the average SMPL head joint position error in millimeters.

\subsection{Body estimation}

In our first set of experiments, we evaluate body estimation from only device SLAM poses, without considering images or hands.
This setting allows us to isolate the advantages of our body motion prior, while directly comparing against methods that do not consider hands.

\subsubsection{Invariant conditioning evaluation}

We begin by evaluating the importance of the spatial and temporal invariance criteria discussed in Section~\ref{sec:invariant_conditioning}.
We do this by comparing five implementations of the conditioning $g$: \textbf{(1)~\texttt{\ours{}}} is the final invariant representation that we propose in Equation~\ref{eq:cond_param}.
\textbf{(2)~\texttt{Absolute+Local Relative}} appends absolute poses with the relative pose deltas written in Equation~\ref{eq:relative_cpf_motion}.
\textbf{(3)~\texttt{Absolute+Global Deltas}} appends absolute poses with relative orientation and the world-frame position deltas used by~\cite{jiang2022avatarposer,castillo2023bodiffusion}.
\textbf{(4)~\texttt{Sequence Canonicalization}} uses the alignment approach implemented by~\cite{li2023egoego}, which violates temporal invariance.
\textbf{(5)~\texttt{Absolute}} naively conditions on absolute poses, which violate spatial invariance.

We train conditional diffusion models with otherwise identical architecture using each parameterization, and then evaluate on the AMASS~\cite{mahmood2019amass} test set.
Metrics and percent differences compared to \ours{} are reported in Table~\ref{table:cond_ablation}.

Overall, we find that the choice of conditioning parameterization makes a dramatic impact on estimation accuracy.
We observe accuracy improve consistently as invariance properties are incorporated into the representation.
Compared to \texttt{\ours{}}, \texttt{Absolute} conditioning increases MPJPE by over 23\% for both shorter (length 32) and longer (length 128) sequences.
Compared to \texttt{\ours{}}, \texttt{SeqCanonical} conditioning increases MPJPE by nearly 18\% for length 32 sequences and 12\% for length 128 sequences.

\input{figtex/quant_body}

\input{figtex/qual_amass_comp_run}

\subsubsection{Comparisons against baselines}

To further study \ours{}'s body estimation quality, we compare against three baselines. %
\textbf{(1)~NoShape.}
First, NoShape refers to a variation of \ours{} that turns off shape estimation, and thus cannot estimate the wearer's height.
\textbf{(2)~EgoEgo.} 
We also compare against the human motion diffusion model from EgoEgo~\cite{li2023egoego}.
This is similar to \ours{}, but considers only the SMPL ``mean'' body shape and uses sequence canonicalized coordinates for conditioning and as model output.
\textbf{(3)~VAE+Opt.}
Finally, we compare against an approach based on the SLAHMR~\cite{ye2023slahmr} framework for human motion estimation from exocentric video.
A key advantage of SLAHMR is that it uses an unconditional motion prior~\cite{rempe2021humor} in an optimization framework.
It can therefore be adapted to new settings without re-training---we keep the same body pose and shape variables as the original pipeline, but replace the exocentric keypoint~\cite{rajasegaran2022tracking} cost with an egocentric CPF pose alignment cost.

Due to differences in problem formulation, many existing methods for egocentric human motion estimation are difficult to directly compare.
This is particularly true when they have different inputs, such as fisheye cameras~\cite{wang2021estimating,tome2019xr,wang2024egocentric}, wrist-mounted sensors~\cite{lee2024mocap}, or handheld controller poses~\cite{castillo2023bodiffusion,jiang2022avatarposer,jiang2023egoposer}.
Additionally, prior works like EgoEgo~\cite{li2023egoego} do not incorporate vision inputs for hand estimation.
For fairness, we restrict all methods in this section to only CPF or head pose as input.

\vspace{0.5em}
\noindent
\textbf{\ours{} improves body motion estimates.}
We report metrics in Table~\ref{table:quant_amass_rich_adt} and visualize example outputs in Figure~\ref{fig:amass_comp_run}.
We find that \ours{} enables significant estimation improvements across all datasets, including accuracy improvements of 20$\sim$30\% over EgoEgo for both shorter and longer evaluation sequences.
We found shape estimation critical for producing metric-scale, grounded estimates of human body motion, with the head aligned to input SLAM poses and the feet planted on the observed ground plane.
This is evident in qualitative results, improved grounding metrics, and in the 6$\sim$7\% MPJPE gap between \ours{} and the NoShape ablation.

\vspace{0.5em}
\noindent
\textbf{VAE optimization converges poorly.}
Optimization-based estimation approaches have been effective for settings with keypoint costs~\cite{rempe2021humor,ye2023slahmr}, but we found convergence difficult in our less constrained setting.
In Table~\ref{table:quant_amass_rich_adt}, we observe poor generalization: VAE+Opt performs competitively on the AMASS test set, but performance deteriorates dramatically when evaluating on RICH or ADT.
VAE+Opt outputs in Figure~\ref{fig:amass_comp_run} also look overly smoothed, without the same expressiveness as the conditional predictions of \ours{} or EgoEgo~\cite{li2023egoego}.
This highlights the advantage of using a conditional diffusion model problem for this estimation problem.

\vspace{0.5em}
\noindent
\textbf{Shape estimation evaluation.}
To better understand the shape estimation characteristics of \ours{}, we compare against against the ``mean'' shape used by EgoEgo and the NoShape ablation.
On the AMASS test set, we find: \ours{} slightly improves overall shape ($19\text{mm}\to18\text{mm}$ mean vertex-to-vertex error) and produces much better height ($52\text{mm}\to32\text{mm}$ mean height error), but is not able to generalize in terms of body weight ($5\text{kg}\to8\text{kg}$ mean weight error).
The body shape is inferred from the wearer's head pose, which intuitively provides strong height constraints but is less correlated with weight.
Accurate height is key for proper scene placement, as reflected by both the MPJPE and GND metrics.

\subsection{Hand estimation}
To evaluate hands estimated by \ours{}, we run HaMeR on the segment of the EgoExo4D~\cite{grauman2023egoexo4d} validation set that is labeled with 3D hand pose keypoints.
We quantitatively compare four hand estimation methods in Table~\ref{table:quant_hands}.
In \textbf{(1)~HaMeR}~\cite{pavlakos2023reconstructing}, we use HaMeR out-of-the-box on undistorted egocentric RGB images.
We do not assume bounding boxes as input; instead, we follow the HaMeR demo code and compute crops using ViTPose~\cite{xu2022vitpose+}.
\textbf{(2)~\ours{}-NoReproj} uses all loss terms except for the reprojection loss (Equation~\ref{eq:reprojection}).
Hand guidance is done directly using the 3D wrist poses predicted by HaMeR.
\textbf{(3)~\ours{}-Mono} is the same as \ours{}-NoReproj, but guides hands using the reprojection loss.
This accounts for the scale ambiguities that are inherent to the single-frame HaMeR estimates.
Finally, \textbf{(4)~\ours{}-Wrist3D} uses both the HaMeR losses and 3D wrist pose losses from Project Aria's Machine Perception Services~\cite{somasundaram2023projectaria}---unlike HaMeR, which assumes monocular input, this uses a pair of SLAM cameras that are unique to Project Aria.
For fairness across settings, we compute metrics only on timesteps where HaMeR estimates are available.

Results are provided in Table~\ref{table:quant_hands}.
While HaMeR's local poses (PA-MPJPE) are slightly better, \ours{}'s hand-body estimation significantly improves how well hands are estimated in the world coordinate system.
Compared to HaMeR, \ours{}-Mono drops MPJPE from $237.90\text{mm}\to131.45\text{mm}$.
Incorporating more accurate wrist pose estimates (\ours{}-Wrist3D) offers a practical solution for further improvements: $131.45\text{mm}\to 60.08\text{mm}$.
Reprojection-based guidance is also important: despite using the same inputs, \ours{}-NoReproj outputs are worse than \ours{}-Mono in both MPJPE and PA-MPJPE.

\begin{figure}[t!]
\centering
\includegraphics[width=\linewidth]{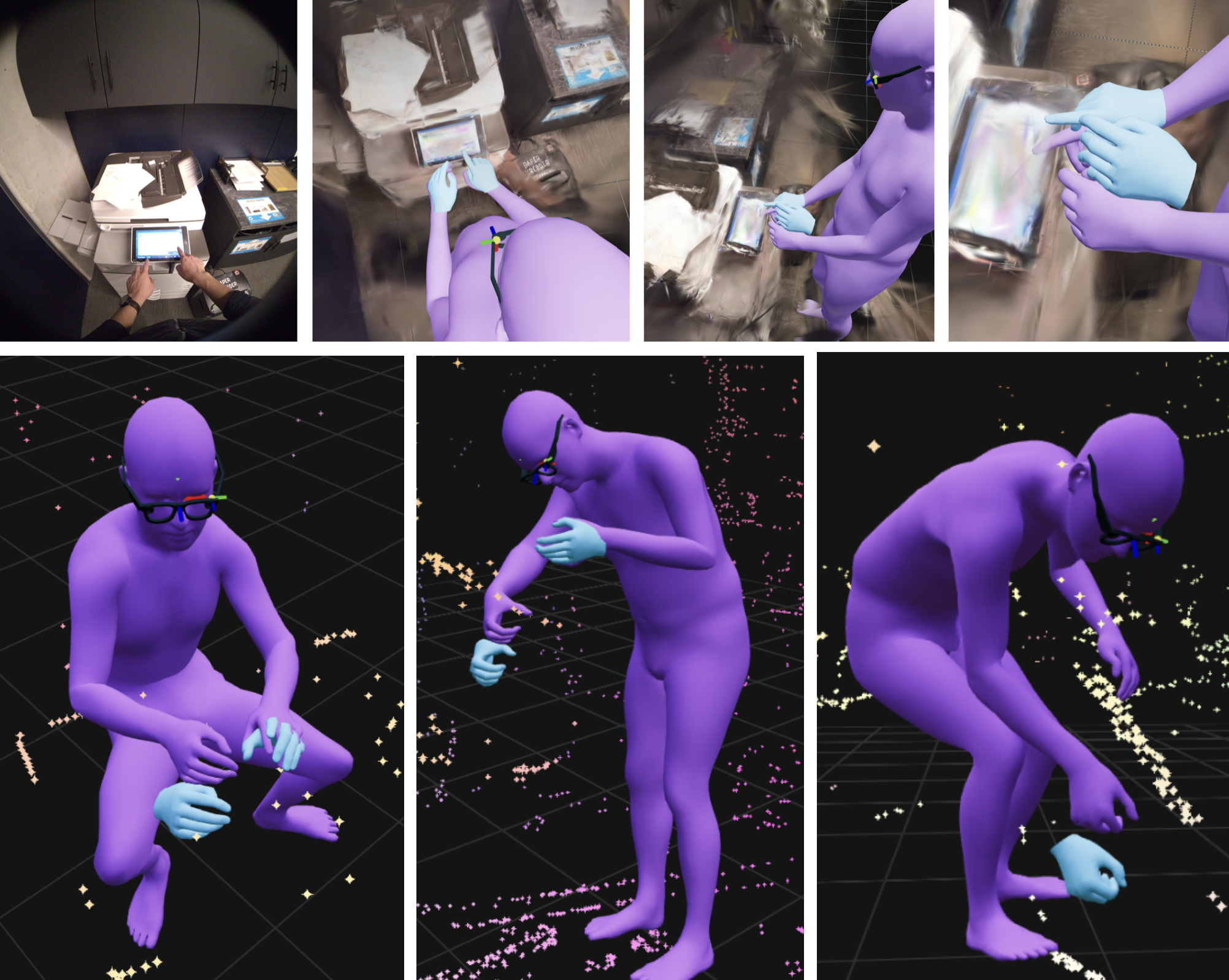}
\caption{
\textbf{Body estimation improves hand estimation.}
We show raw outputs from HaMeR~\cite{pavlakos2023reconstructing} in blue and hand-body estimations from \ours{} in purple.
\textit{Top:} improved scene interaction during touchscreen operation with \ours{}-Mono.
We know a priori that the fingers are contacting the screen in this sequence.
\textit{Bottom:} qualitative examples from EgoExo~\cite{grauman2023egoexo4d} evaluation, showing the differences between monocular hands and \ours{}-Wrist3D estimates.
}
\label{fig:qual_hands}
\end{figure}

Qualitatively, we observed that high hand estimation errors in naive monocular estimation with HaMeR are explained by a combination of detection failures and monocular ambiguities.
Even when detections succeed, the scale and distance of monocular HaMeR estimates are often incorrect or flicker in between frames.
Incorporating these hands via guidance with our diffusion motion prior encourages final outputs that obey the kinematic and smoothness constraints imposed by plausible body motion---we provide examples of HaMeR estimates rendered jointly with \ours{} outputs in Figure~\ref{fig:qual_hands}. %

\input{figtex/quant_hands}
\label{table:hands_quantitative}

%% file: figtex/cond_compare.tex
\newcommand{\compareyes}{\makebox[1.25em][c]{{\color{ForestGreen} \ding{52}}}}
\newcommand{\compareno}{\makebox[1.25em][c]{{\color{BrickRed} \ding{55}}}}
\newcommand{\comparepartially}{\makebox[1.25em][c]{{\color{Dandelion} \textsf{\textbf{P}}}}}

\begin{table*}[t]
\begin{center}
\resizebox{\linewidth}{!}{%
\begin{tabular}{llcccccc}
\toprule
Conditioning & Seqlen & Invariance~\ref{inv:spatial_invariance} /~\ref{inv:temporal_invariance} & MPJPE $\downarrow$ & \% Diff & PA-MPJPE $\downarrow$ & \% Diff & GND $\uparrow$\\
\cmidrule(r){1-1} \cmidrule(lr){2-2} \cmidrule(lr){3-3} \cmidrule(){4-5} \cmidrule(){6-7} \cmidrule(){8-8}
\texttt{\ours{}} (Eq.~\ref{eq:cond_param}) & 32 & \compareyes\quad/\quad\compareyes & \cellcolor{yellow!30}129.8\scriptsize{$\pm$1.1} & \cellcolor{yellow!30}--- & \cellcolor{yellow!30}109.8\scriptsize{$\pm$1.1} & \cellcolor{yellow!30}--- & \cellcolor{yellow!30}0.98\scriptsize{$\pm$0.00}\\

\texttt{Absolute+Local Relative} & 32 & \comparepartially\quad/\quad\compareyes & \cellcolor{orange!30}133.0\scriptsize{$\pm$1.1} & \cellcolor{orange!30}2.4\% & \cellcolor{orange!30}113.6\scriptsize{$\pm$1.2} & \cellcolor{orange!30}3.4\% & \cellcolor{orange!30}0.95\scriptsize{$\pm$0.00}\\

\texttt{Absolute+Global Deltas}~\cite{castillo2023bodiffusion,jiang2022avatarposer} & 32 & \compareno\quad/\quad\compareyes & \cellcolor{red!30}136.2\scriptsize{$\pm$1.1} & \cellcolor{red!30}4.9\% & \cellcolor{red!30}118.3\scriptsize{$\pm$1.2} & \cellcolor{red!30}7.7\% & \cellcolor{red!30}0.93\scriptsize{$\pm$0.01}\\

\texttt{Sequence Canonicalization}~\cite{li2023egoego} & 32 & \compareyes\quad/\quad\compareno & 153.1\scriptsize{$\pm$1.5} & 17.9\% & 128.7\scriptsize{$\pm$1.5} & 17.1\% & 0.76\scriptsize{$\pm$0.01}\\

\texttt{Absolute} & 32 & \compareno\quad/\quad\compareyes & 159.9\scriptsize{$\pm$1.2} & 23.2\% & 141.0\scriptsize{$\pm$1.3} & 28.4\% & 0.89\scriptsize{$\pm$0.01}\\

\cmidrule(r){1-1} \cmidrule(lr){2-2} \cmidrule(lr){3-3} \cmidrule(){4-5} \cmidrule(){6-7} \cmidrule(){8-8}
\texttt{\ours{}} (Eq.~\ref{eq:cond_param}) & 128 & \compareyes\quad/\quad\compareyes & \cellcolor{yellow!30}119.7\scriptsize{$\pm$1.3} & \cellcolor{yellow!30}--- & \cellcolor{yellow!30}101.1\scriptsize{$\pm$1.3} & \cellcolor{yellow!30}--- & \cellcolor{yellow!30}1.00\scriptsize{$\pm$0.00}\\

\texttt{Absolute+Local Relative} & 128 & \comparepartially\quad/\quad\compareyes & \cellcolor{orange!30}124.5\scriptsize{$\pm$1.3} & \cellcolor{orange!30}4.0\% & \cellcolor{orange!30}104.9\scriptsize{$\pm$1.4} & \cellcolor{orange!30}3.8\% & \cellcolor{orange!30}1.00\scriptsize{$\pm$0.00}\\

\texttt{Absolute+Global Deltas}~\cite{castillo2023bodiffusion,jiang2022avatarposer} & 128 & \compareno\quad/\quad\compareyes & \cellcolor{red!30}127.4\scriptsize{$\pm$1.3} & \cellcolor{red!30}6.4\% & \cellcolor{red!30}109.8\scriptsize{$\pm$1.4} & \cellcolor{red!30}8.6\% & \cellcolor{red!30}0.99\scriptsize{$\pm$0.00}\\

\texttt{Sequence Canonicalization}~\cite{li2023egoego} & 128 & \compareyes\quad/\quad\compareno & 134.0\scriptsize{$\pm$1.8} & 11.9\% & 112.1\scriptsize{$\pm$1.6} & 10.9\% & 0.88\scriptsize{$\pm$0.02}\\

\texttt{Absolute} & 128 & \compareno\quad/\quad\compareyes & 148.3\scriptsize{$\pm$1.5} & 23.9\% & 131.2\scriptsize{$\pm$1.6} & 29.8\% & 0.96\scriptsize{$\pm$0.01}\\

\bottomrule
\end{tabular}
}

\end{center}

\vspace{-1em}
\caption{
    \textbf{Motion prior conditioning comparison.}
    We train and evaluate otherwise identical models using four conditioning parameterizations on AMASS~\cite{mahmood2019amass} test set sequences, using sequences of length 32 and 128.
    Parameterizations vary in their spatial (\ref{inv:spatial_invariance}) and temporal (\ref{inv:temporal_invariance}) invariance properties, which we loosely classify as following completely (\compareyes), partially (\comparepartially), or not at all (\compareno).
    The conditioning parameterization used by \ours{} reduces errors by almost 18\% compared to the sequence canonicalization approach used by the most relevant related work~\cite{li2023egoego}.
}
\vspace{-0.5em}
\label{table:cond_ablation}
\end{table*}

%% file: figtex/quant_body.tex
\begin{table}[t]
	\begin{center}
\resizebox{\linewidth}{!}{%
\begin{tabular}{llccccc}
\toprule
\multicolumn{4}{l}{\textbf{AMASS~\cite{mahmood2019amass}}}\\
\midrule
Method & Seq & MPJPE $\downarrow$ & PA-MPJPE $\downarrow$ & GND $\uparrow$ & $\mathbf{T}_\text{head}$ $\downarrow$ \\
\cmidrule(lr){1-1} \cmidrule(lr){2-2} \cmidrule(lr){3-6}
\ours{} & 32 & \cellcolor{yellow!30}129.8\scriptsize{$\pm$1.1} & \cellcolor{yellow!30}109.8\scriptsize{$\pm$1.1} & \cellcolor{yellow!30}0.98\scriptsize{$\pm$0.00} & \cellcolor{yellow!30}6.4\scriptsize{$\pm$0.1}\\
NoShape & 32 & \cellcolor{orange!30}138.1\scriptsize{$\pm$1.1} & \cellcolor{orange!30}118.8\scriptsize{$\pm$1.1} & \cellcolor{orange!30}0.94\scriptsize{$\pm$0.01} & \cellcolor{orange!30}44.7\scriptsize{$\pm$0.4}\\
EgoEgo & 32 & \cellcolor{red!30}184.0\scriptsize{$\pm$1.5} & \cellcolor{red!30}158.6\scriptsize{$\pm$1.6} & \cellcolor{red!30}0.81\scriptsize{$\pm$0.01} & \cellcolor{red!30}45.2\scriptsize{$\pm$1.0}\\
VAE+Opt & 32 & 199.5\scriptsize{$\pm$1.3} & 191.4\scriptsize{$\pm$1.4} & 0.49\scriptsize{$\pm$0.01} & 78.0\scriptsize{$\pm$1.5}\\
\cmidrule(lr){1-1} \cmidrule(lr){2-2} \cmidrule(lr){3-6}
\ours{} & 128 & \cellcolor{yellow!30}119.7\scriptsize{$\pm$1.3} & \cellcolor{yellow!30}101.1\scriptsize{$\pm$1.3} & \cellcolor{yellow!30}1.0\scriptsize{$\pm$0.00} & \cellcolor{yellow!30}6.2\scriptsize{$\pm$0.1}\\
NoShape & 128 & \cellcolor{orange!30}128.1\scriptsize{$\pm$1.3} & \cellcolor{orange!30}110.3\scriptsize{$\pm$1.4} & \cellcolor{orange!30}0.98\scriptsize{$\pm$0.01} & \cellcolor{orange!30}44.6\scriptsize{$\pm$0.7}\\
EgoEgo & 128 & \cellcolor{red!30}167.4\scriptsize{$\pm$2.1} & \cellcolor{red!30}145.8\scriptsize{$\pm$2.0} & \cellcolor{red!30}0.92\scriptsize{$\pm$0.01} & \cellcolor{red!30}54.9\scriptsize{$\pm$1.9}\\
VAE+Opt & 128 & 205.3\scriptsize{$\pm$2.6} & 192.3\scriptsize{$\pm$2.8} & 0.75\scriptsize{$\pm$0.02} & 67.8\scriptsize{$\pm$3.1}\\
\midrule
\multicolumn{4}{l}{\textbf{RICH~\cite{Huang:CVPR:2022}}}\\
\midrule
Method & Seq & MPJPE $\downarrow$ & PA-MPJPE $\downarrow$ & GND $\uparrow$ & $\mathbf{T}_\text{head}$ $\downarrow$ \\
\cmidrule(lr){1-1} \cmidrule(lr){2-2} \cmidrule(lr){3-6}
\ours{} & 32 & \cellcolor{yellow!30}193.7\scriptsize{$\pm$3.4} & \cellcolor{yellow!30}174.8\scriptsize{$\pm$3.6} & \cellcolor{yellow!30}0.95\scriptsize{$\pm$0.01} & \cellcolor{yellow!30}8.8\scriptsize{$\pm$0.2}\\
NoShape & 32 & \cellcolor{orange!30}200.9\scriptsize{$\pm$3.3} & \cellcolor{orange!30}183.3\scriptsize{$\pm$3.6} & \cellcolor{orange!30}0.73\scriptsize{$\pm$0.02} & \cellcolor{orange!30}44.9\scriptsize{$\pm$0.9}\\
EgoEgo & 32 & \cellcolor{red!30}215.4\scriptsize{$\pm$3.9} & \cellcolor{red!30}192.9\scriptsize{$\pm$4.0} & \cellcolor{red!30}0.73\scriptsize{$\pm$0.02} & \cellcolor{red!30}56.2\scriptsize{$\pm$2.9}\\
VAE+Opt & 32 & 352.0\scriptsize{$\pm$6.7} & 354.8\scriptsize{$\pm$6.5} & 0.59\scriptsize{$\pm$0.02} & 319.3\scriptsize{$\pm$11.6}\\
\cmidrule(lr){1-1} \cmidrule(lr){2-2} \cmidrule(lr){3-6}
\ours{} & 128 & \cellcolor{yellow!30}176.2\scriptsize{$\pm$5.6} & \cellcolor{yellow!30}160.1\scriptsize{$\pm$5.9} & \cellcolor{yellow!30}0.96\scriptsize{$\pm$0.02} & \cellcolor{yellow!30}8.9\scriptsize{$\pm$0.3}\\
NoShape & 128 & \cellcolor{orange!30}185.7\scriptsize{$\pm$5.5} & \cellcolor{orange!30}169.9\scriptsize{$\pm$5.8} & \cellcolor{red!30}0.82\scriptsize{$\pm$0.03} & \cellcolor{orange!30}45.8\scriptsize{$\pm$1.6}\\
EgoEgo & 128 & \cellcolor{red!30}207.8\scriptsize{$\pm$6.9} & \cellcolor{red!30}187.8\scriptsize{$\pm$6.8} & \cellcolor{orange!30}0.88\scriptsize{$\pm$0.03} & \cellcolor{red!30}66.5\scriptsize{$\pm$5.4}\\
VAE+Opt & 128 & 319.8\scriptsize{$\pm$10.1} & 323.8\scriptsize{$\pm$10.5} & 0.75\scriptsize{$\pm$0.04} & 274.4\scriptsize{$\pm$17.6}\\
\midrule
\multicolumn{4}{l}{\textbf{Aria Digital Twins~\cite{pan2023aria}}}\\
\midrule
Method & Seq & MPJPE $\downarrow$ & PA-MPJPE $\downarrow$ & GND $\uparrow$ & $\mathbf{T}_\text{head}$ $\downarrow$ \\
\cmidrule(lr){1-1} \cmidrule(lr){2-2} \cmidrule(lr){3-6}
\ours{} & 32 & \cellcolor{yellow!30}173.5\scriptsize{$\pm$1.1} & \cellcolor{yellow!30}146.1\scriptsize{$\pm$1.1} & \cellcolor{orange!30}0.88\scriptsize{$\pm$0.01} & -\\
NoShape & 32 & \cellcolor{orange!30}178.5\scriptsize{$\pm$1.1} & \cellcolor{orange!30}153.0\scriptsize{$\pm$1.1} & \cellcolor{yellow!30}0.89\scriptsize{$\pm$0.01} & -\\
EgoEgo & 32 & \cellcolor{red!30}212.5\scriptsize{$\pm$1.4} & \cellcolor{red!30}181.3\scriptsize{$\pm$1.6} & \cellcolor{red!30}0.64\scriptsize{$\pm$0.01} & -\\
VAE+Opt & 32 & 284.9\scriptsize{$\pm$1.6} & 283.9\scriptsize{$\pm$1.9} & 0.63\scriptsize{$\pm$0.01} & -\\
\cmidrule(lr){1-1} \cmidrule(lr){2-2} \cmidrule(lr){3-6}
\ours{} & 128 & \cellcolor{yellow!30}155.1\scriptsize{$\pm$1.6} & \cellcolor{yellow!30}129.3\scriptsize{$\pm$1.6} & \cellcolor{orange!30}0.94\scriptsize{$\pm$0.01} & -\\
NoShape & 128 & \cellcolor{orange!30}163.7\scriptsize{$\pm$1.6} & \cellcolor{orange!30}140.0\scriptsize{$\pm$1.6} & \cellcolor{yellow!30}0.96\scriptsize{$\pm$0.01} & -\\
EgoEgo & 128 & \cellcolor{red!30}182.6\scriptsize{$\pm$2.3} & \cellcolor{red!30}153.9\scriptsize{$\pm$2.6} & \cellcolor{red!30}0.73\scriptsize{$\pm$0.02} & -\\
VAE+Opt & 128 & 290.8\scriptsize{$\pm$3.8} & 282.5\scriptsize{$\pm$4.4} & 0.7\scriptsize{$\pm$0.02} & -\\
\bottomrule
\end{tabular}
}
	\end{center}
        \vspace{-0.3em}
	\caption{
            \textbf{Body estimation performance, compared against a baseline without shape prediction, EgoEgo~\cite{li2023egoego}, and VAE+Opt~\cite{rempe2021humor,ye2023slahmr}.}
            We exclude the $\textbf{T}_\text{head}$ metric for ADT because the Biomech57 head joints used by ADT are not directly comparable to the SMPL-H head joints used by our model.
        }
	\label{table:quant_amass_rich_adt}
    \vspace{-1em}
\end{table}

%% file: figtex/qual_amass_comp_run.tex
\begin{figure}[t!]
\centering

\begin{subfigure}{0.235\textwidth}
  \centering
  \includegraphics[width=\linewidth]{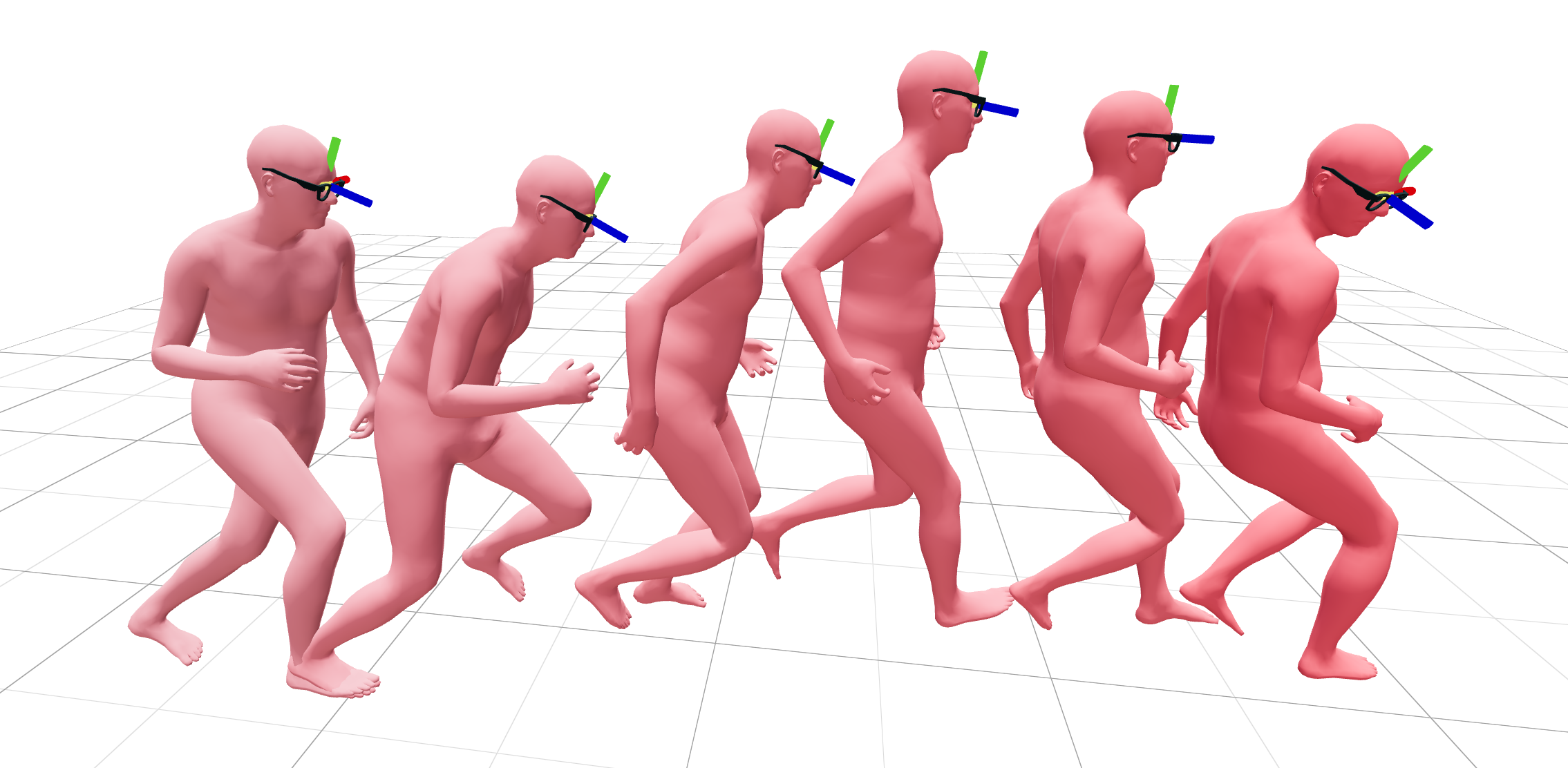}
  \caption{\centering Ground-truth}
\end{subfigure}
\hfill
\begin{subfigure}{0.235\textwidth}
  \centering
  \includegraphics[width=\linewidth]{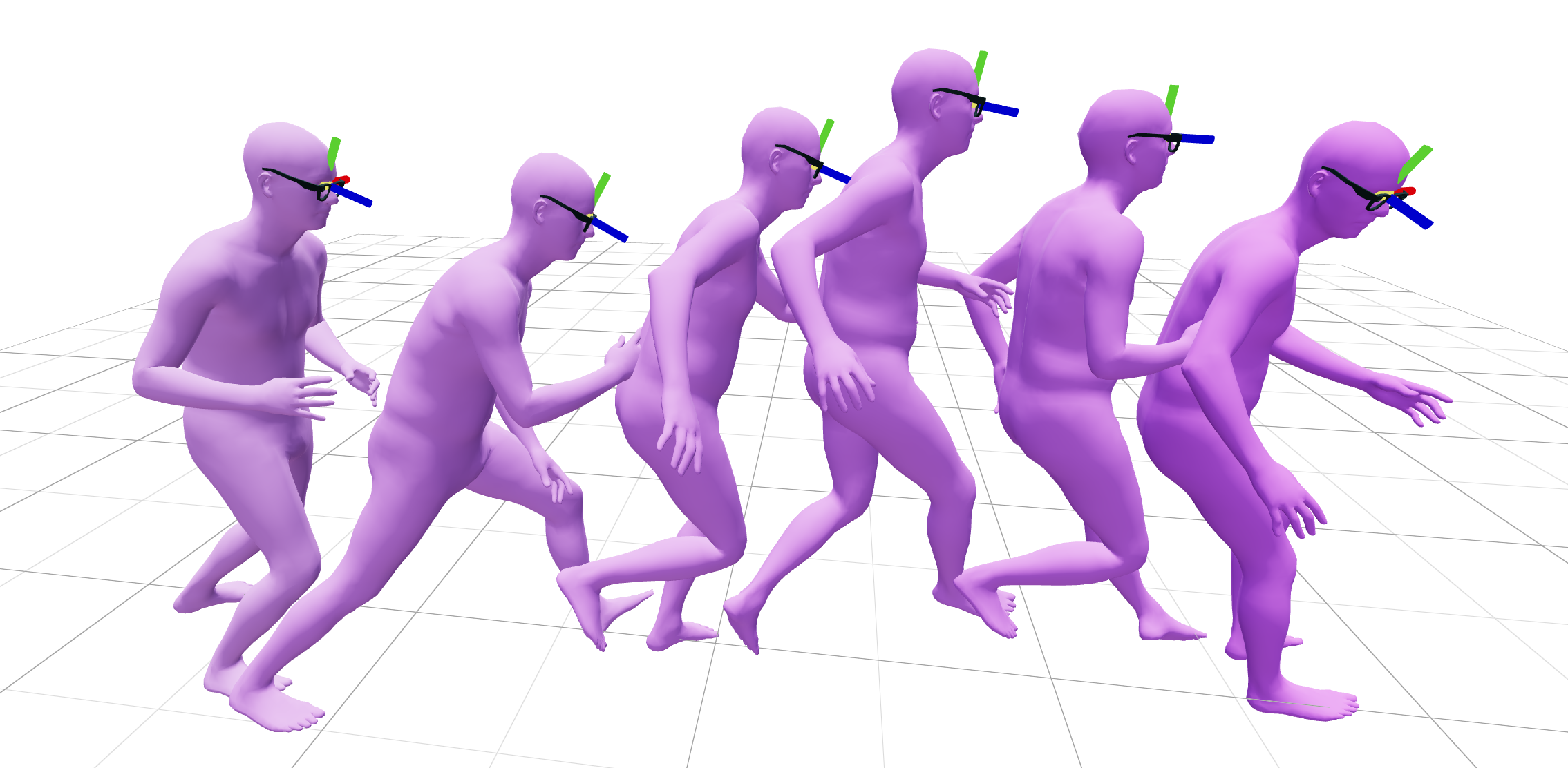}
  \vspace{-0.5em}
  \caption{\ours{}}
\end{subfigure}

\begin{subfigure}{0.235\textwidth}
  \centering
  \includegraphics[width=\linewidth]{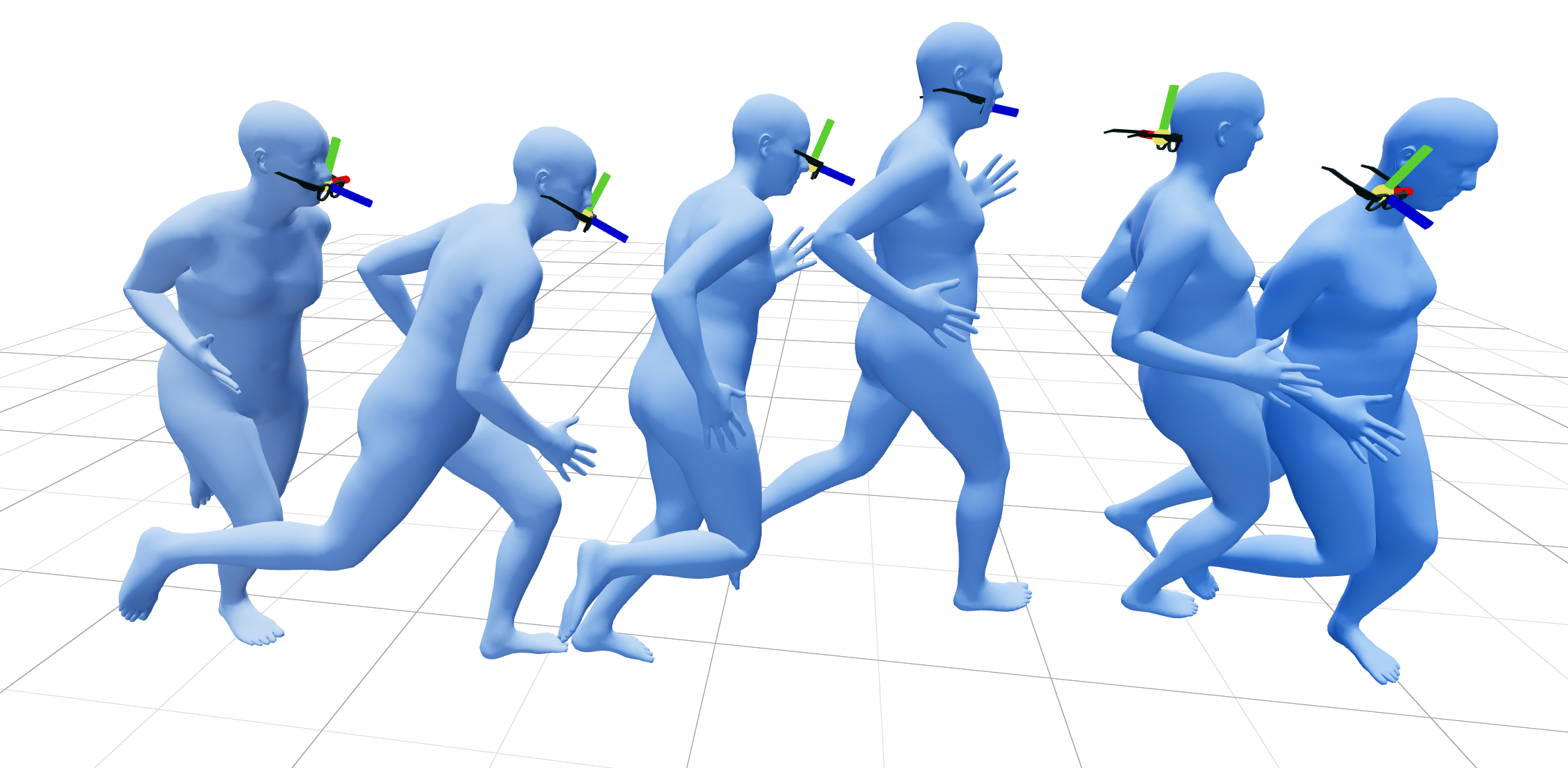}
  \vspace{-0.5em}
  \caption{EgoEgo~\cite{li2023egoego}}
\end{subfigure}
\hfill
\begin{subfigure}{0.235\textwidth}
  \centering
  \includegraphics[width=\linewidth]{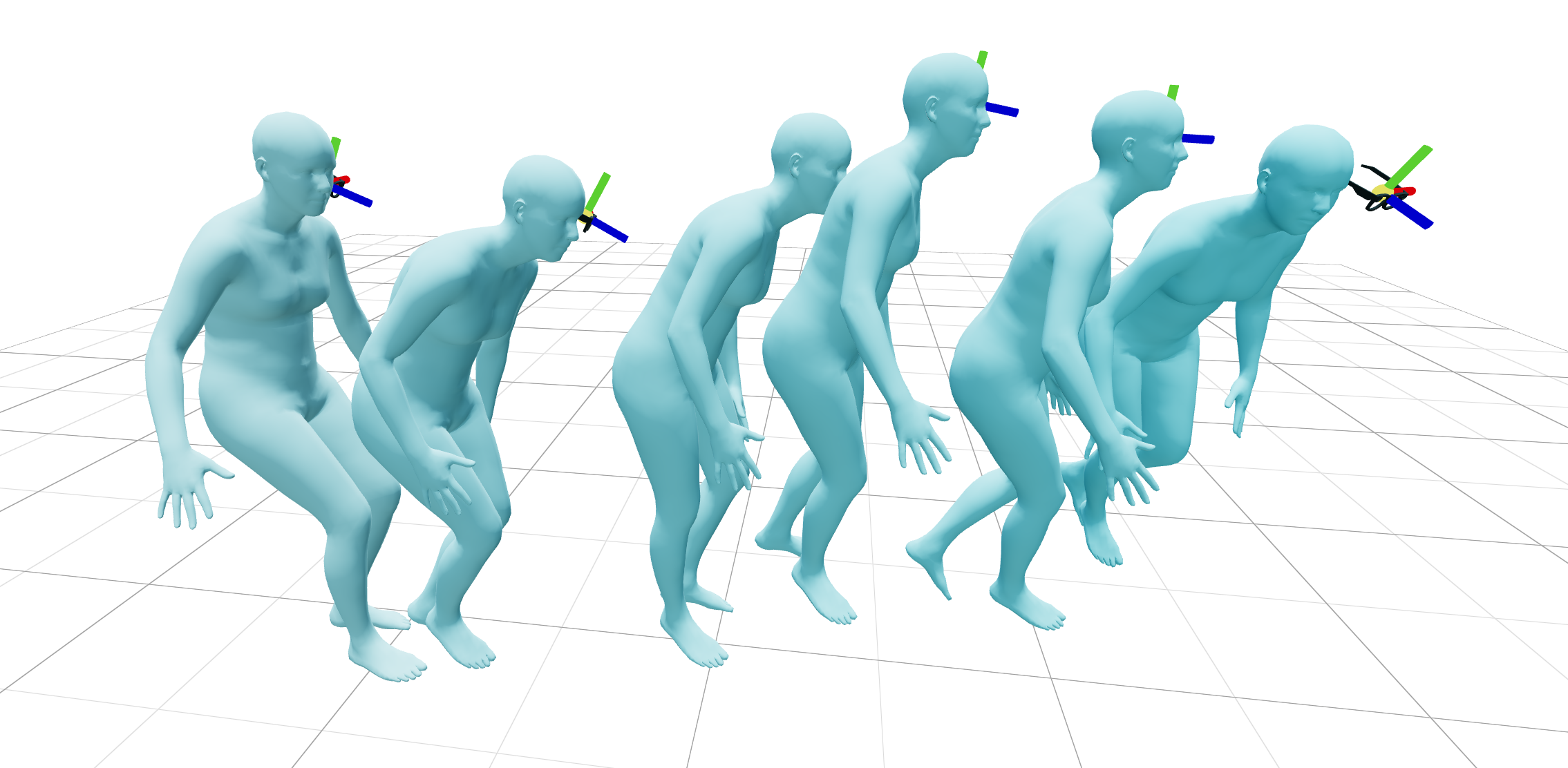}
  \vspace{-0.5em}
  \caption{VAE+Opt}
\end{subfigure}

\caption{
\textbf{Egocentric human motion estimation for a running sequence.}
We show the ground-truth, an output from \ours{}, and outputs from two baselines.
The glasses CAD model is placed at the conditioning transformation $\textbf{T}_{\text{world},\text{cpf}}$.
}
\label{fig:amass_comp_run}
\vspace{-0.5em}
\end{figure}

%% file: figtex/quant_hands.tex
\newcommand{\bestcell}[1]{\cellcolor{yellow!30}{#1}}
\newcommand{\secondbestcell}[1]{\cellcolor{orange!30}{#1}}
\newcommand{\thirdbestcell}[1]{\cellcolor{red!30}{#1}}
\begin{table}[t]
	\begin{center}
 \begin{subtable}[t]{0.48\textwidth}
 \centering
\begin{tabular}{lcc}
\toprule
Method & MPJPE $\downarrow$ & PA-MPJPE $\downarrow$ \\
\cmidrule(r){1-1} \cmidrule(){2-3}
HaMeR & 237.90\scriptsize{$\pm$1.89} & \bestcell{13.04\scriptsize{$\pm$1.89}} \\
\cmidrule(r){1-1} \cmidrule(){2-3}
\ours{}-NoReproj & \thirdbestcell{143.20\scriptsize{$\pm$0.42}} & 14.75\scriptsize{$\pm$0.42} \\
\ours{}-Mono & \secondbestcell{131.45\scriptsize{$\pm$0.39}} & \thirdbestcell{14.71\scriptsize{$\pm$0.39}} \\
\ours{}-Wrist3D & \bestcell{60.08\scriptsize{$\pm$0.26}} & \secondbestcell{14.38\scriptsize{$\pm$0.26}} \\

\bottomrule
\end{tabular}

    \end{subtable}

    \end{center}
    \caption{\textbf{Hand estimation errors in millimeters.}
            \ours{}'s hand-body estimation can constrain and resolve ambiguities in noisy outputs from HaMeR, which we observe can reduce MPJPE for hands by over 40\%.
            }
    \label{table:quant_hands}
    \vspace{-1em}
\end{table}

%% file: sec/5_conclusion.tex
\section{Dicussion}

\noindent
\textbf{Limitations and future work.}
While the core contributions of \ours{} are general, the current implementation of our system has a few limitations that we hope to explore in future work.
First, diffusion model guidance is a test-time optimization process that depends on hyperparameters and incurs a runtime cost.
In the future, it may be possible to bootstrap using outputs from our model to train a feedforward model that avoids this step.
Success for hand guidance also still depends on reasonable monocular hand estimates.
Estimation can therefore fail as a result of errors like left/right flipping or spurious detections. %
Finally, we assume flat floors.
This is in part because our training data~\cite{mahmood2019amass} includes floor planes but no detailed scene geometry.
As a result, our method will fail in settings like hills or staircases.
In the future, we hope to extend our insights to data with more detailed scene information, which concurrent work has highlighted the usefulness of in informing human body estimation~\cite{guzov2024hmd}.

\vspace{0.5em}
\noindent
\textbf{Conclusion.}
We presented \ours{}, a system for estimating human motion using sensors from head-mounted devices.
\ours{} jointly estimates human body pose, height, and hand parameters from only egocentric SLAM poses and images.
Results highlight the importance of spatial and temporal invariance in conditioning for this problem, while demonstrating how estimated bodies can be used to improve hand estimation.

%% file: sec/X_suppl.tex
\setcounter{page}{1}
\maketitlesupplementary

\renewcommand{\thetable}{A.\arabic{table}}
\setcounter{table}{0}  %
\renewcommand{\thesection}{A.\arabic{section}}
\setcounter{section}{0}  %
\renewcommand{\thefigure}{A.\arabic{figure}}
\setcounter{figure}{0}  %

\section{Invariant Conditioning Visualization}
\label{app:conditioning_vis}

As we observe in Table~\ref{table:cond_ablation}, naively training a model using absolute head poses results in poor estimation performance.
The absence of spatial invariance (Invariance~\ref{inv:spatial_invariance}) explains this result.
To visualize this, we show in Figure~\ref{fig:cond_abs_poses} two renders of the same human motion trajectory.
The second render has the same local body motion as the first, but with the world frame re-defined:

\begin{figure}[h!]
    \centering
    \begin{subfigure}{0.26\textwidth}
      \centering
      \includegraphics[width=\linewidth]{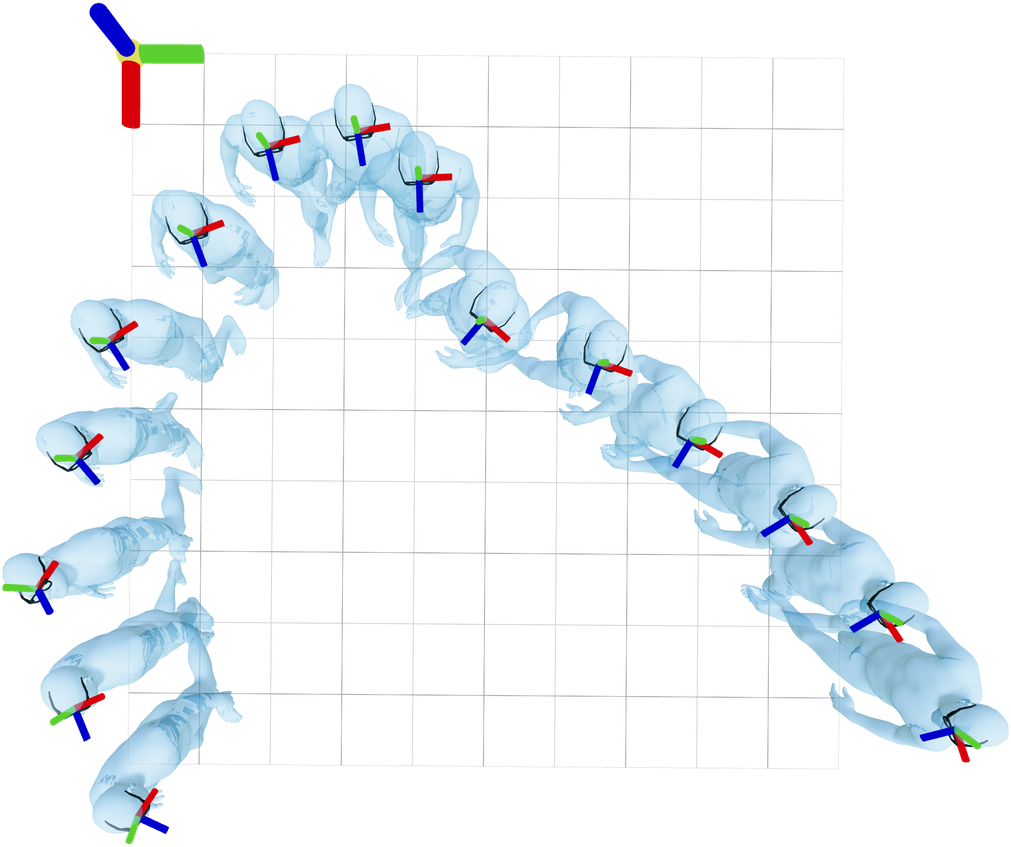}
      \caption{Before}
    \end{subfigure}
    \hfill
    \begin{subfigure}{0.2\textwidth}
      \centering
      \includegraphics[width=\linewidth]{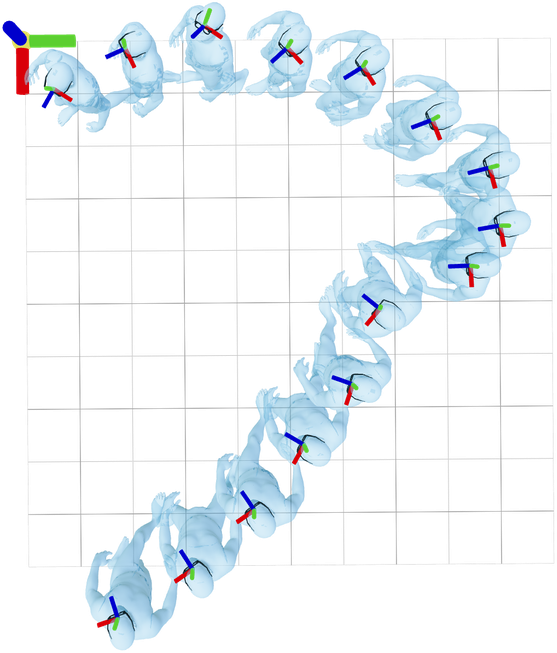}
      \caption{After}
    \end{subfigure}
    
    \caption{
    \textbf{Absolute head pose visualization for a single human motion trajectory, before and after re-defining the world frame.}
    } 
    \label{fig:cond_abs_poses}
\end{figure}

Because the world frame location is arbitrarily defined, naively conditioning on these poses hinders generalization.
Works like EgoPoser~\cite{jiang2023egoposer} have made similar observations.

To fix this, prior works have preprocessed sequences by aligning them to a canonical coordinate frame located at the first timestep of each sequence~\cite{li2023egoego,rempe2021humor,guzov2024hmd}.
However, we observe that this is flawed from the perspective of temporal invariance (invariance~\ref{inv:temporal_invariance}).
To visualize this, we render in Figure~\ref{fig:canonical_slices} two temporal slices of the same body motion, with one slice starting from the beginning of the motion and another starting from the middle:

\begin{figure}[h!]
    \centering
    \begin{subfigure}{0.2\textwidth}
      \centering
      \includegraphics[width=\linewidth]{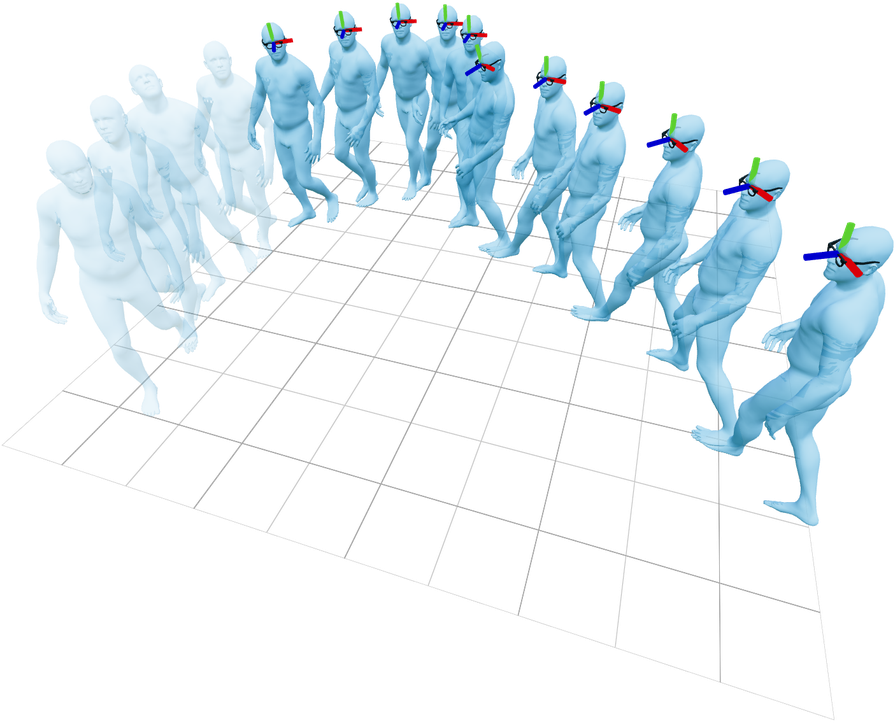}
      \caption{First slice}
    \end{subfigure}
    \hfill
    \begin{subfigure}{0.24\textwidth}
      \centering
      \includegraphics[width=\linewidth]{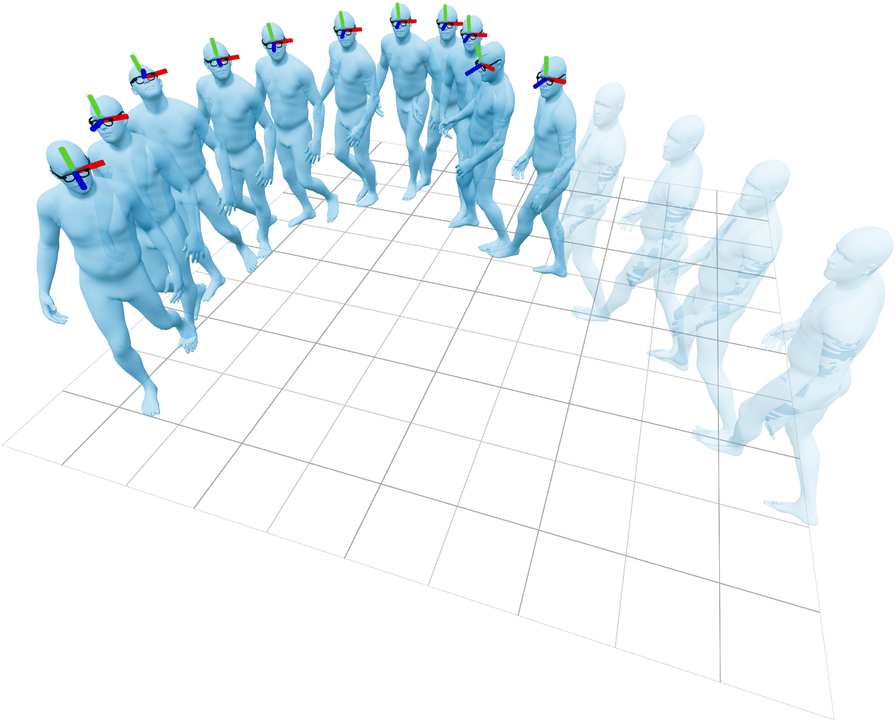}
      \caption{Second slice}
    \end{subfigure}
    \caption{
        \textbf{Two slices of the same human motion trajectory.}
    } 
    \label{fig:canonical_slices}
\end{figure}

\begin{figure}[h!]
    \centering
    \begin{subfigure}{0.2\textwidth}
      \centering
      \includegraphics[width=\linewidth]{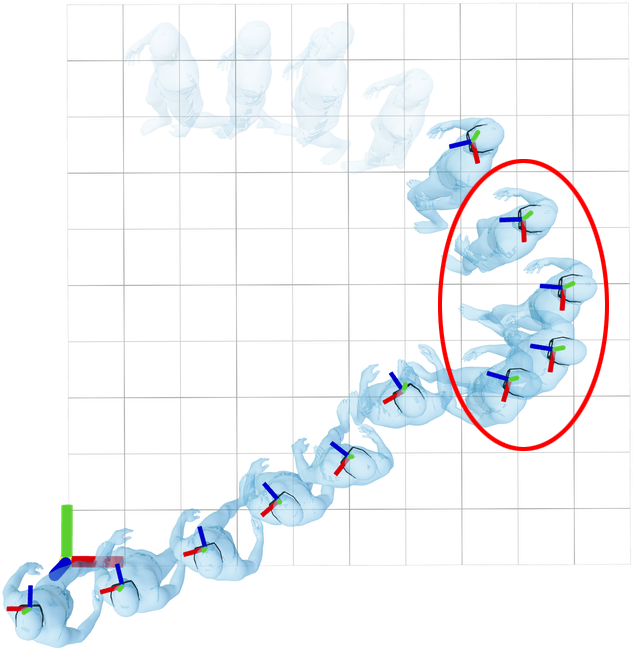}
      \caption{First slice}
    \end{subfigure}
    \hfill
    \begin{subfigure}{0.24\textwidth}
      \centering
      \includegraphics[width=\linewidth]{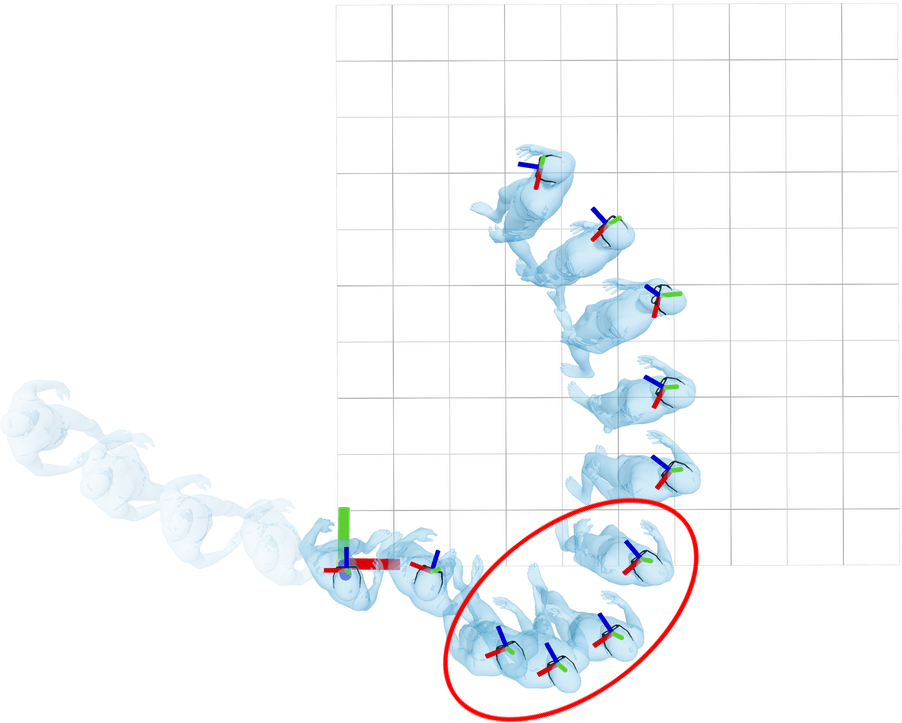}
      \caption{Second slice}
    \end{subfigure}
    \caption{
        \textbf{Poses canonicalized by aligning the first timestep.}
    } 
    \label{fig:canonical_top_circled}
\end{figure}

Next, we consider how the head pose trajectories for each of these slices would look if they were canonicalized by aligning the first timestep.
We visualize the resulting head pose trajectories in Figure~\ref{fig:canonical_top_circled}.
Circled in {\color{red} red} are four timesteps that are shared between the two slices.
Notice that head poses from canonicalized sequences can still differ significantly, even for the same body motion.

\begin{figure}[h!]
  \centering
  \includegraphics[width=0.5\textwidth]{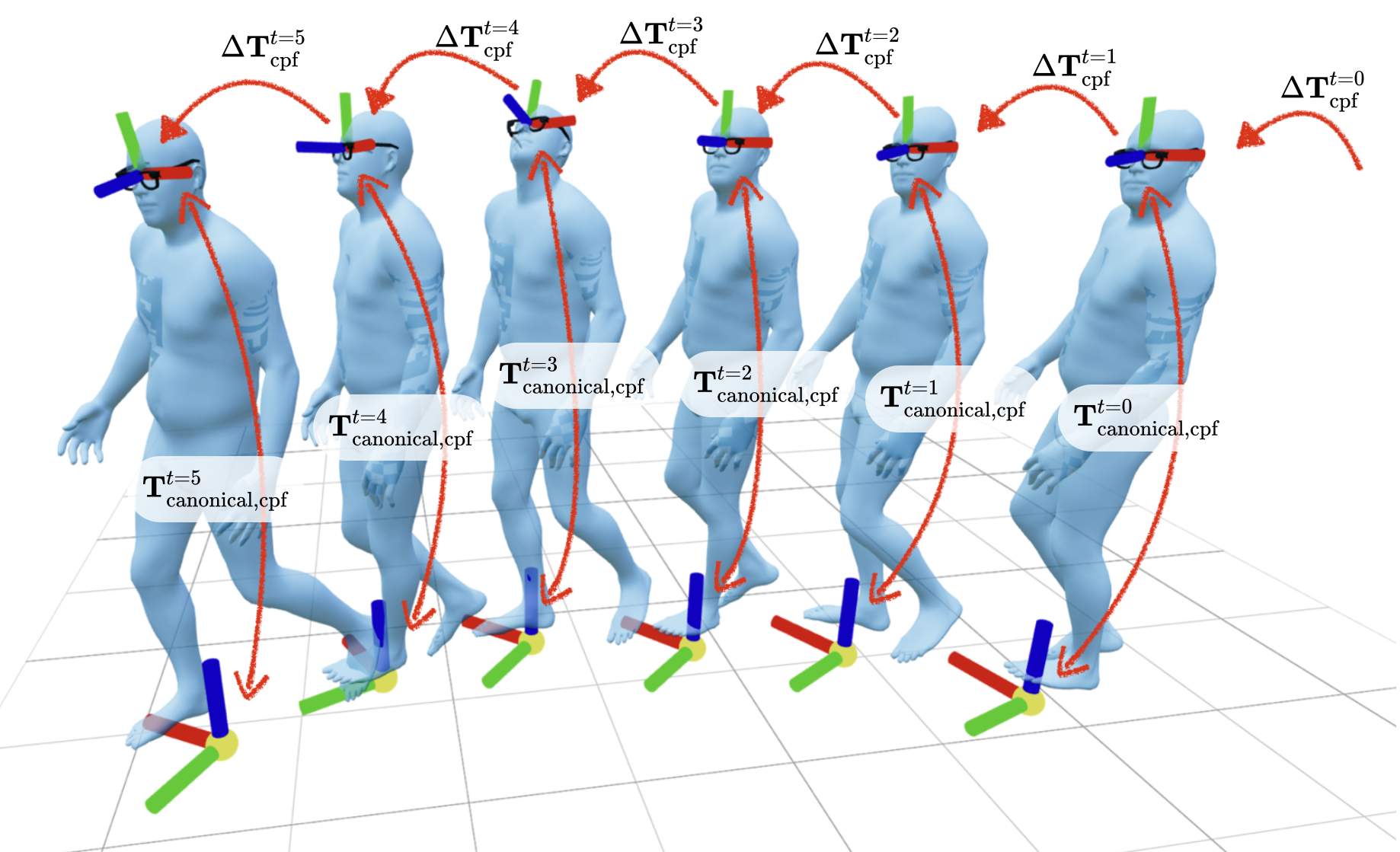}
  \caption{
    \textbf{Transformations that make up the invariant conditioning used by \ours{}.}
  }
  \label{fig:conditioning_annotated}
\end{figure}

To achieve both Invariances~\ref{inv:spatial_invariance} and \ref{inv:temporal_invariance}, \ours{}'s invariant conditioning paramterization proposes an alternative way to canonicalize head poses.
Instead of defining a single canonical coordinate frame for each temporal window, we define a canonical coordinate frame at \textit{every} timestep.
The resulting representation couples relative CPF motion $\Delta\mathbf{T}_{\text{cpf}}^{t}$ with per-timestep canonicalized pose $\mathbf{T}_{\text{canonical, cpf}}^{t}$.
These transformations are visualized in Figure~\ref{fig:conditioning_annotated}.
Notice that the transformations that make up this conditioning approach are invariant both to the world coordinate system and to choices in temporal windowing.
This enables significant improvements in estimation accuracy (Table~\ref{table:cond_ablation}).

\section{Ancillary Results}

\subsection{Sequence length evaluation}
At test time, \ours{} follows MultiDiffusion~\cite{bar2023multidiffusion} for extrapolating to arbitary sequence lengths.
To validate this choice, we filter out test sequences shorter than 256 frames and then evaluate both \ours{} and EgoEgo~\cite{li2023egoego} with subsequences of length 32, 128, and 256.
We report MPJPE metrics on these sequences in Table~\ref{tab:long_sequence_tab}.
Both \ours{} and EgoEgo include windowing strategies for handling longer sequences; unlike prior work, however, we find that accuracy improves even after test set sequence lengths surpass the training set sequence length.
\input{figtex/quant_longer_seqs}

\subsection{Additional qualitative results}

We provide additional qualitative results for the body motion prior in Figure~\ref{fig:amass_comp_squat}.
\ours{} estimates have the head aligned exactly to input observations and the feet planted realistically on the floor.

\input{figtex/qual_amass_comp_squat}

\section{Implementation Details}

\subsection{Network architecture}

\ours{} uses a transformer~\cite{vaswani2017attention} architecture with rotary positional embeddings~\cite{su2024roformer} for its denoising model $\mu_\theta(\vec{x}_n, \vec{c}, n)$.
Sampling is performed by denoising all timesteps within a temporal window in parallel: we do not sample autoregressively and therefore do not use causal masking.
\textit{Encoder details:} latent encodings $\vec{z}_c$ are computed as output from conditioning sequences $\vec{c}$ as input using six transformer blocks, each containing a self-attention layer followed by a 2-layer MLP.
\textit{Decoder details:} the denoised output is computed using six additional transformer blocks that take $\vec{x}_n$ as input, while conditioning on $\vec{z}_c$ via cross-attention.
All hidden dimensions are set to 512.

\textbf{Runtime.} For a length-128 sequence, each forward pass through \ours{}'s denoising network takes 0.05 seconds on a single RTX 4090.
Because we use DDIM~\cite{song2020denoising} for sampling, the number of denoising steps for each sample can be chosen to make tradeoffs between sample quality and speed.
All experiments in our paper use 30 DDIM steps.

\subsection{Guidance optimizer}
For guidance, we use a Levenberg-Marquardt optimizer implemented in JAX~\cite{jax2018github}.
Levenberg-Marquardt is an iterative nonlinear least squares algorithm, which requires solving a linearized subproblem at each timestep.
We compute the Jacobians needed for this as block-sparse matrices for efficiency, and solve the resulting linear subproblems using a Conjugate Gradient optimizer.

\begin{figure}[h!]
    
    \begin{subfigure}[b]{\linewidth}
        \centering
        \includegraphics[width=.9\linewidth]{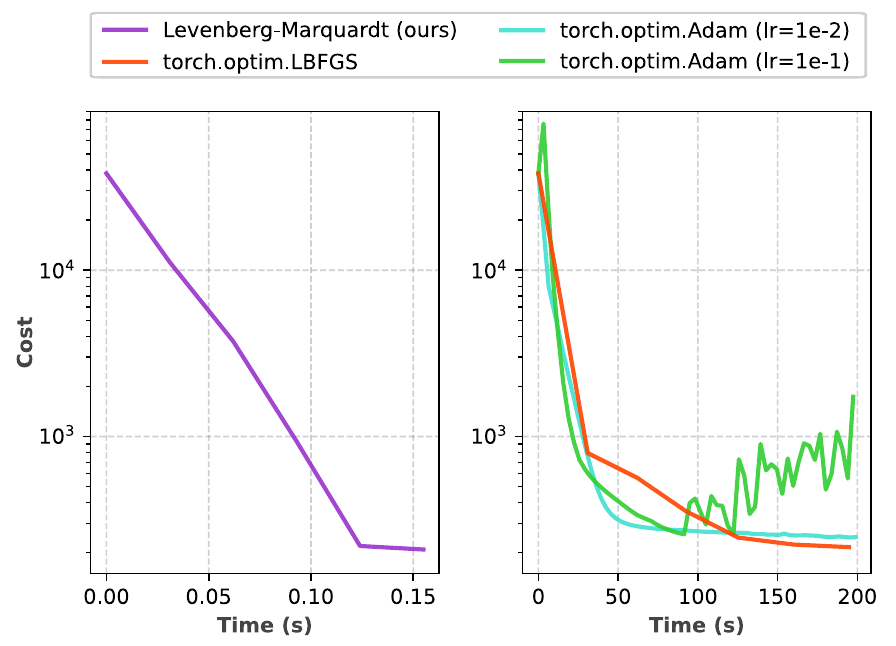}
        \caption{
            \textbf{Costs over time.}
             LM converges significantly faster than off-the-shelf PyTorch optimizers for guidance optimization.
        }
        \vspace{.5em}
        \label{fig:optimizer_timings_plot}
    \end{subfigure}

    \vspace{1em}
    
    \begin{subfigure}[b]{\linewidth}
        \centering

        \scalebox{.8}{
            \begin{tabular}{lr}
                \toprule
                Optimizer & Final Cost \\
                \midrule
                Levenberg-Marquardt (ours) & 209.05 \\
                torch.optim.LBFGS & 215.96 \\
                torch.optim.Adam (lr=1e-2) & 248.01 \\
                torch.optim.Adam (lr=1e-1) & 1733.30 \\
                \bottomrule
            \end{tabular}
        }
        \vspace{.5em}
        \caption{
            \textbf{Final costs.}
            We report the final cost for each method in the plot above.
        }
        \label{fig:optimizer_timings_table}
    \end{subfigure}
    
    \caption{
        \textbf{Comparing guidance optimizers.}
    }
    \label{fig:optimizer_timings}
\end{figure}

\textbf{Runtime.} The guidance optimizer converges in 0.15$\sim$0.2 seconds on an RTX 4090.
We compare our LM optimizer against off-the-shelf PyTorch optimizers in Figure~\ref{fig:optimizer_timings}.

\subsection{Floor height estimation}
\label{app:floor_height}

\begin{figure}[t!]
  \centering
  \includegraphics[width=\linewidth]{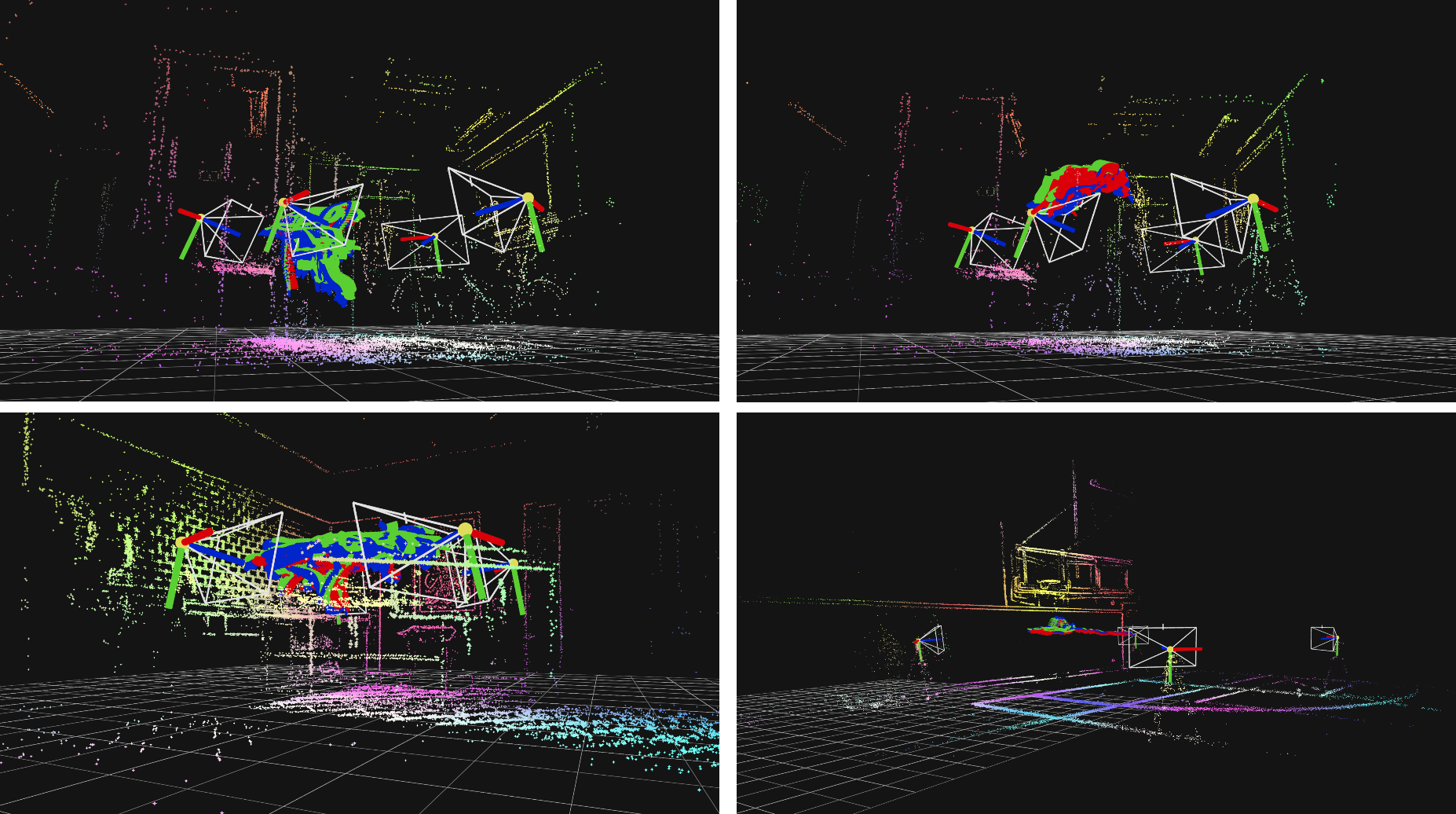}
  \caption{
    \textbf{Floor height examples.}
    Point cloud-derived floor height examples on the EgoExo4D dataset.
  }
  \label{fig:floor_height}
\end{figure}

One requirements of \ours{} is SLAM poses that can be situated relative to the floor.
While floor heights are provided in our training data, they are not directly available on real-world data.
We found that a RANSAC-based algorithm works well on real-world data from Project Aria~\cite{pan2023aria}.
We filter SLAM points by confidence, then use RANSAC to find a z-value with that best fits a plane.
Example floor plane outputs using scenes from the EgoExo4D~\cite{grauman2023egoexo4d} dataset are shown in Figure~\ref{fig:floor_height}.

\subsection{Biomech57 evaluation details}

The majority of our evaluation data (AMASS~\cite{mahmood2019amass} and RICH~\cite{Huang:CVPR:2022}) is provided directly using SMPL conventions.
Because \ours{} outputs SMPL-H parameters, this makes computation of joint error metrics straightforward.

The one exception is the Aria Digital Twins dataset~\cite{pan2023aria}, which we use for quantitative body metrics.
Each device wearer in the Aria Digital Twins dataset is recorded via an Optitrack motion capture system, which records 57 joint locations (30 hand joints, 27 body joints) following the Biomech57 joint template.
To evaluate our method on ADT, we match and compare the common major joints between the two templates. %
We manually corresponded each of the 57 joints between Biomech57 and the standard SMPL-H joint conventions.
While the majority of these have 1:1 correspondences---feet, knees, hips, shoulders, elbows, wrist, and finger joints, for example, are consistently defined---we mask out others like the head and collar bone joints that are misaligned.

%% file: figtex/quant_longer_seqs.tex
\begin{table}[t]
    \centering
    \resizebox{.5\columnwidth}{!}{
    \begin{tabular}{cccc}
    \toprule
   Seqlen  & 32 & 128 & 256 \\
     \midrule
     EgoAllo & 149.3 & 130.3 & 127.9\\
     EgoEgo & 187.7 & 173.8 & 184.3\\
    \bottomrule
    \end{tabular}
    }
    \vspace{-0.3em}
    \caption{
    \textbf{Effect of sequence length on MPJPE (mm).}
    \ours{} is trained with sequences of length 128. EgoEgo~\cite{li2023egoego} is trained with sequences of length 140.
    }
    \label{tab:long_sequence_tab}
\end{table}

%% file: figtex/qual_amass_comp_squat.tex
\begin{figure}[t]
\centering

\begin{subfigure}{0.38\textwidth}
  \centering
  \includegraphics[width=\linewidth]{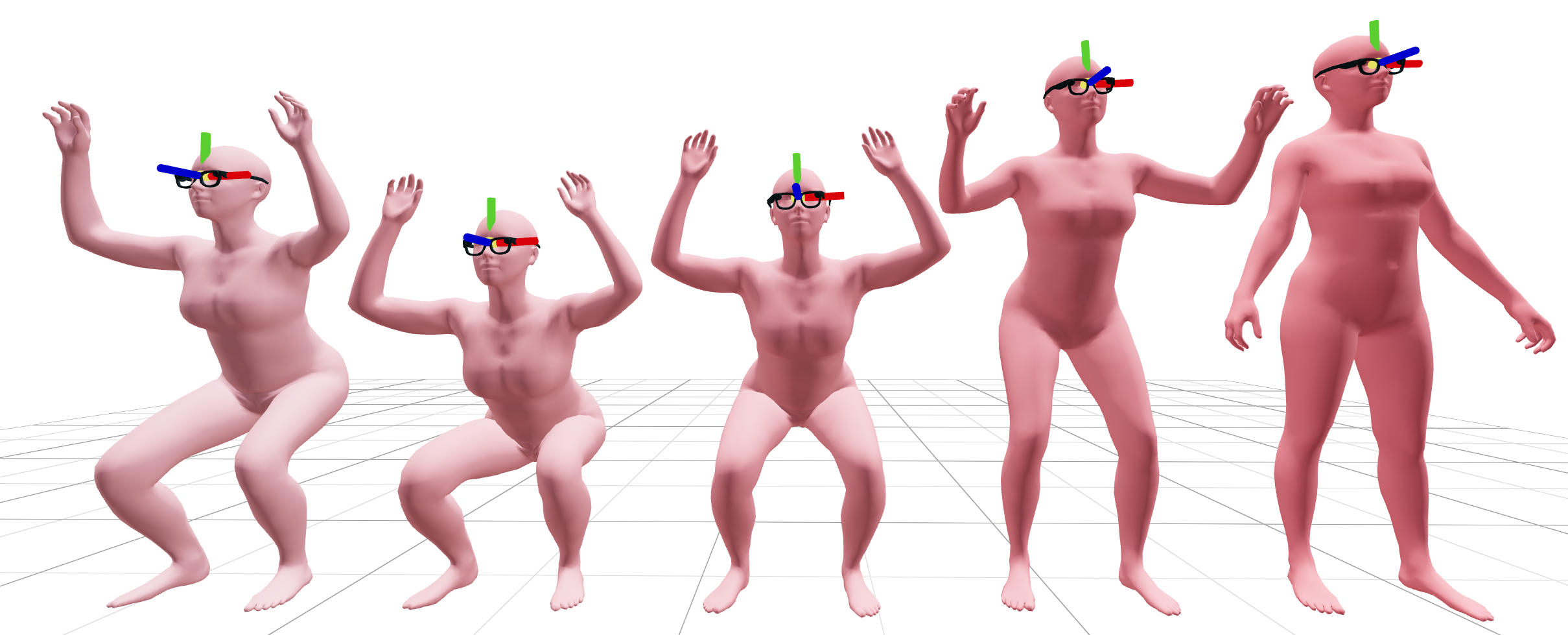}
  \caption{Ground-truth}
\end{subfigure}
\hfill
\begin{subfigure}{0.38\textwidth}
  \centering
  \includegraphics[width=\linewidth]{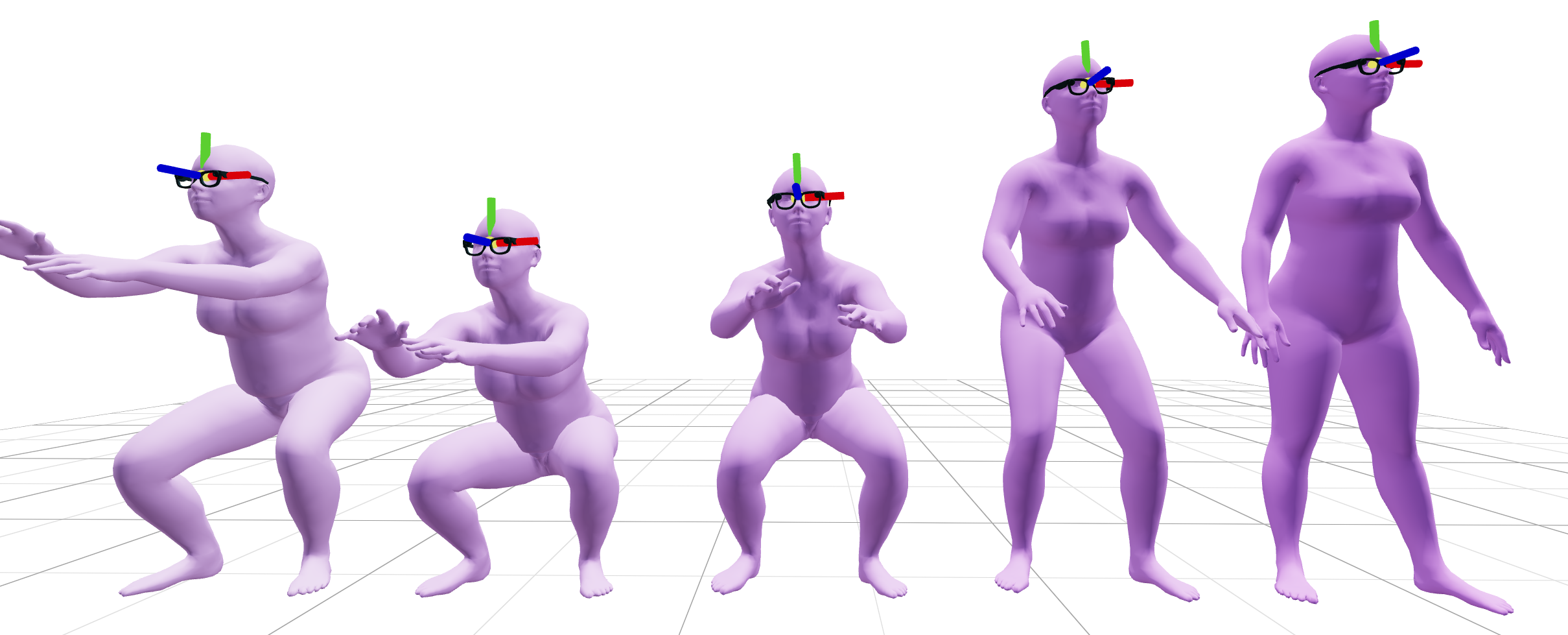}
  \caption{\ours{}}
\end{subfigure}

\begin{subfigure}{0.38\textwidth}
  \centering
  \includegraphics[width=\linewidth]{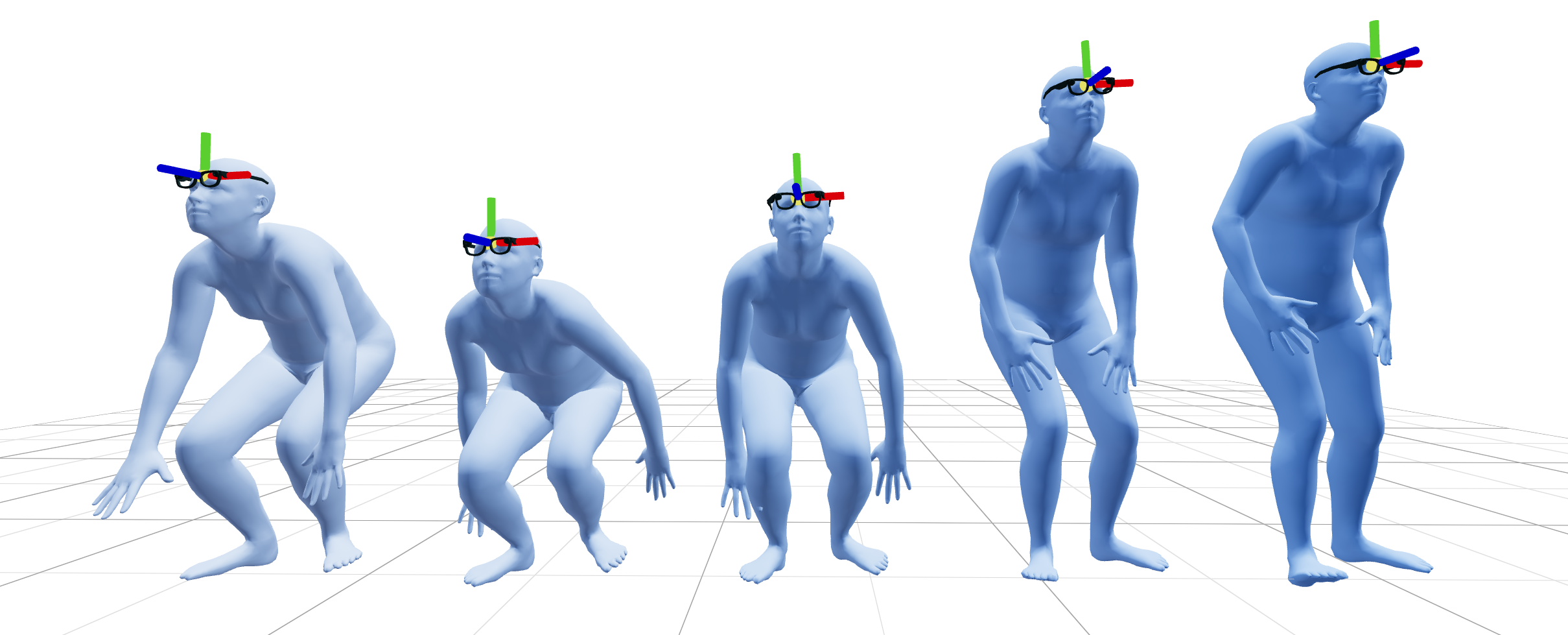}
  \caption{EgoEgo~\cite{li2023egoego}}
\end{subfigure}
\hfill
\begin{subfigure}{0.38\textwidth}
  \centering
  \includegraphics[width=\linewidth]{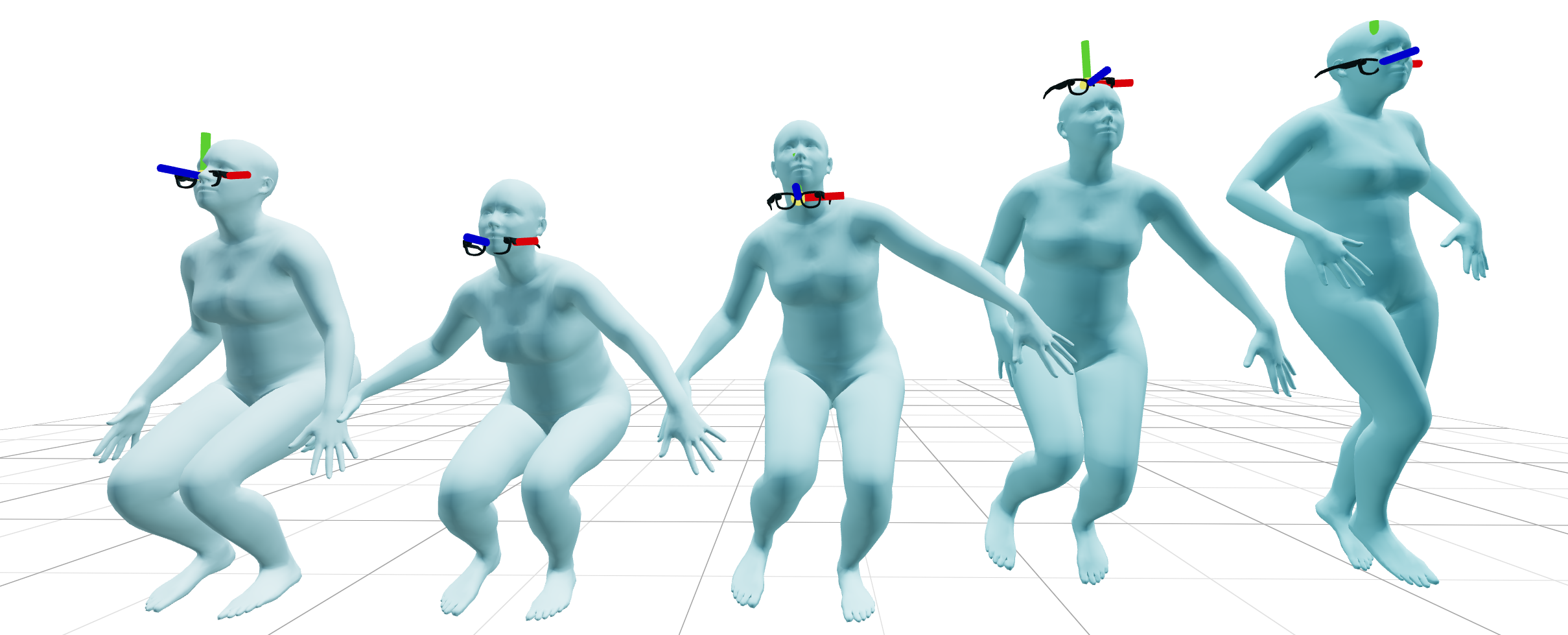}
  \caption{VAE+Opt}
\end{subfigure}

\caption{
\textbf{Head pose-conditioned motion prior results for a squatting sequence.}
Spatial shifts are used to visualize different timesteps within the sequence.
Hand observations are not used.
}

\label{fig:amass_comp_squat}
\end{figure}